\documentclass{article}

\PassOptionsToPackage{numbers}{natbib}
 \usepackage[preprint]{neurips_2026}

\usepackage[utf8]{inputenc} % allow utf-8 input
\usepackage[T1]{fontenc}    % use 8-bit T1 fonts
\usepackage{hyperref}       % hyperlinks
\usepackage{url}            % simple URL typesetting
\usepackage{booktabs}       % professional-quality tables
\usepackage{amsfonts}       % blackboard math symbols
\usepackage{nicefrac}       % compact symbols for 1/2, etc.
\usepackage{microtype}      % microtypography
\usepackage{xcolor}         % colors
\usepackage{graphicx}
\usepackage{amsmath}
\usepackage{enumitem}    % No Indent Enumeration
\usepackage{titlesec}
\titlespacing*{\paragraph}{0pt}{0pt}{1em}
\usepackage{wrapfig}
\usepackage{algorithm}
\usepackage{algpseudocode}

\usepackage{multirow}
\usepackage{array}
\usepackage{adjustbox} 

\newcommand{\up}{$\uparrow$}
\newcommand{\dn}{$\downarrow$}

\title{Thinking in Boxes: \\ 3D Editing in Real Images Made Easy}

\author{
  \begin{tabular}[t]{c}
    Pradhaan S Bhat$^{1}$\thanks{Equal Contribution} \quad
    Naveen Chandra R$^{1}$\footnotemark[1] \quad
    Rishubh Parihar$^{1}$ \quad 
    Vaibhav Vavilala$^{2}$ \quad \\
    R. Venkatesh Babu$^{1}$  \quad
    D.A. Forsyth$^{3}$ \quad 
    Anand Bhattad$^{4}$ 
    \end{tabular}%
  \quad
  \and
  \begin{tabular}[t]{c} 
    $^1$Indian Institute of Science \quad 
    $^2$Apple \quad 
    $^3$UIUC \quad
    $^4$Johns Hopkins University 
  \end{tabular}% 
  \vspace{5pt} \\
  \textbf{Project Page:} \url{https://thinking-in-boxes.github.io/}
}

\begin{document}
\maketitle

\vspace{-16pt}
\begin{figure*}[h]
    \centering
    \includegraphics[width=1.0\linewidth]{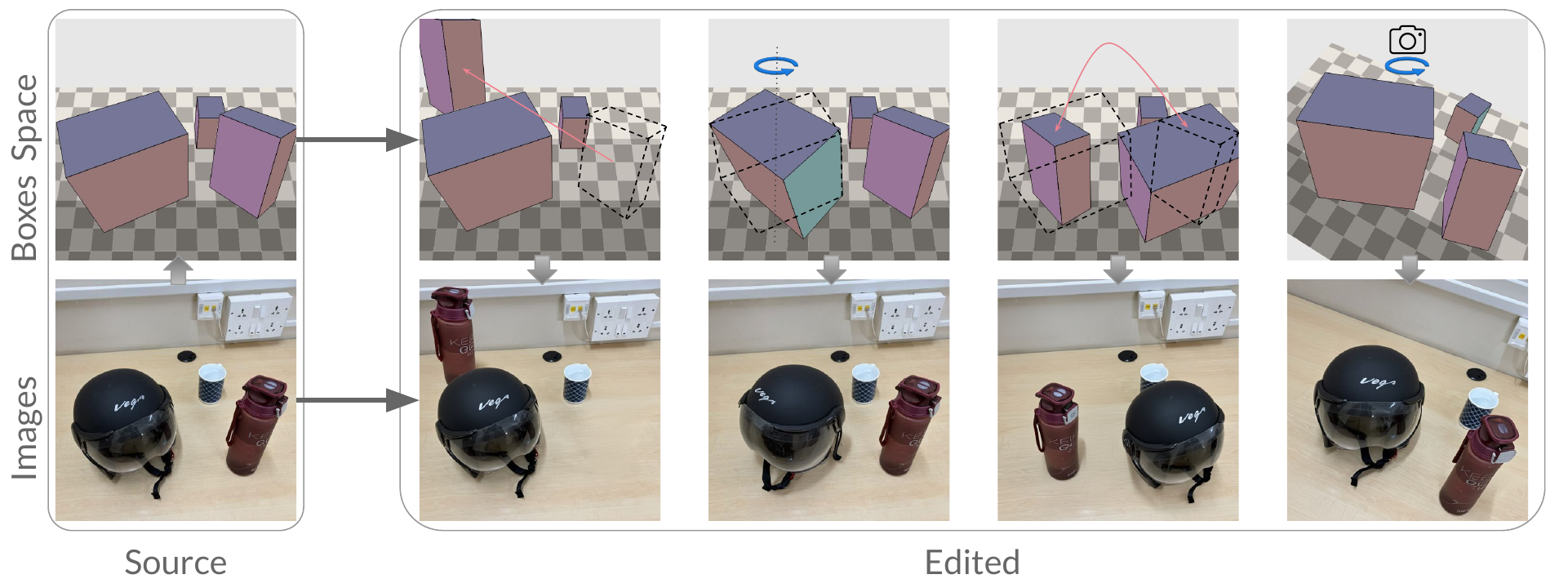}
    \vspace{-15pt}
    \caption{\textbf{Thinking in boxes.} Given a single source image (bottom left), the user fits 3D boxes around objects of interest, anchored on a depth-aligned floor (top left). Editing the boxes drives the corresponding edit in the image: from left to right, translation, rotation, combined translation and rotation, and a camera viewpoint change. The bottles' and helmets' appearance is preserved across all four edits, including regions of the object not visible in the source — note the back of the bottle and side view of the helmet revealed under rotation. Note that all qualitative figures are real images.
    }
    \label{fig:teaser}
\end{figure*}

\begin{abstract}
Text and 2D-conditioning interfaces provide weak, ambiguous control
over spatial transformations in image editing -- particularly under
large object motions and camera changes. Prior work has used 3D
primitives such as boxes, but only as loose conditioning signals
indicating approximate object location rather than specifying the
transformation. We instead use 3D boxes as structured specifications:
the user provides the input and output boxes of the edit, casting
editing as a well-posed geometry problem. This ``thinking in boxes''
interface, where each box face is color-coded to convey 3D orientation, gives precise control over translation, rotation, scaling,
and viewpoint changes in real images while preserving scene and
object identity, and recovering previously unseen object regions. To
ground transformations in scene appearance, we introduce a
depth-aligned planar floor as a global reference frame, shaded with
depth-aware cues. Conditioned on this structure, an image generator
produces consistent results under large transformations. Trained in
two stages -- on synthetic multi-object scenes and a small set of
real-world videos from Objectron -- the system
generalizes to complex, in-the-wild real images. Our method operates directly on real photographs and substantially
outperforms recent state-of-the-art methods on large 3D edits.
\end{abstract}

\section{Introduction}
\label{sec:intro}
We present a system that lets a user manipulate a simple, abstracted representation of the objects in an image, and turns those manipulations into edits of the image itself. The user specifies an edit by placing a 3D box around an object in the input image and a second box where they want the object to end up; a learned procedure maps this pair of boxes into the internal representation of a generator, producing the edited image. The same procedure handles translation, rotation, occlusion, and viewpoint changes and is learned once for all manipulations -- not retrained or optimized per image.  

We call this interface ``\textbf{thinking in boxes}'' (Fig.~\ref{fig:teaser}): the box pair implicitly defines the desired translation, rotation, and scale in 3D, while its projection tells the generator which regions become newly visible and how the object's silhouette should change under perspective. To ground edits in scene appearance, we further introduce a depth-aligned planar floor, shaded with depth-aware cues to provide relative spatial positioning of objects and background.

Our method is trained in two stages: first on synthetic edit sequences with multiple objects per scene, then on a small set of real-world scenes from Objectron~\cite{objectron}. It generalizes to complex, real-world images, significantly outperforming state-of-the-art methods on large spatial edits and successfully recovering previously occluded object regions never seen in the input. We find the depth-aligned floor to be extremely important: it provides the global reference frame that lets the same procedure handle object motion and camera motion within a single coherent representation, rather than treating them as separate problems or using specialized video models. 

Large-scale text-conditioned image generators~\citep{flux,qwen-image} have become a powerful prior for image editing, controlled through
instruction-based prompts~\citep{flux-kontext} or 2D signals like
bounding boxes~\citep{zhang2023adding} and drag points~\citep{shi2024dragdiffusion}.
These interfaces are coarse for spatial edits. Text prompts cannot
specify how far an object should move~\cite{michel2023object} and a 2D box cannot
disambiguate between translation, rotation, and camera motion. For
spatial edits -- moving objects through 3D space -- text and 2D interfaces access
only a fraction of the degrees of freedom the task requires.

Several recent methods aim to fill this gap, each with significant
tradeoffs in the choice of input representation. One line uses
per-object latent representations~\citep{neural-assets,neural-usd} that
work well in restricted domains -- driving scenes, household objects --
but transfer poorly to in-the-wild photos. A second line uses
depth as the 3D representation, lifting diffusion activations or
attention maps to 3D via the depth proxy~\citep{geodiffuser,
diffusion-handles,loosecontrol}; this supports 3D edits on real images
without retraining, but the depth-only representation is brittle under
large transformations and significant disocclusion, and the methods
require per-image inversion or optimization. A third line uses
detailed 3D primitives -- meshes or convex blocks~\citep{gbw,3d-fixup} --
applies the edit in image or depth space, and uses diffusion as a
final cleanup pass; this paradigm has so far been shown on
generated images rather than real photographs~\citep{gbw}. We propose
a different choice. Boxes at the object level and a depth-aligned
floor at the scene level forms a minimal representation --
the user only places and moves boxes -- and structurally complete:
boxes encode object pose and visibility, the floor encodes the scene
geometry and disambiguates the object from the camera motion. Given that
representation, a generator trained once on the source--target pair
renders large 3D edits, including substantial rotations and
disocclusions, on in-the-wild photographs.
\section{Related Work}
\label{sec:related_works}

\paragraph{3D Control in Diffusion Models.} 
Recently, several methods~\cite{seethrough3d,compass-control,loosecontrol} have been proposed to enable 3D control in text-to-image generation. LooseControl~\cite{loosecontrol} leverages the depth of 3D bounding boxes and trains a ControlNet~\cite{controlnet} to condition for 3D-aware generation. Build-A-Scene~\cite{build-a-scene} iteratively applies this process to perform interactive 3D Layout Control. Another major direction of work is to condition the text-to-image diffusion model on individual 3D properties, such as camera control~\cite{cheng2024learning} or object orientations~\cite{compass-control,cd360,minorigen}. These methods encode the 3D representation into a text embedding space to condition the generator. More recently, SeeThrough3D~\cite{seethrough3d} demonstrated occlusion-aware 3D layout control by designing a translucent 3D box layout for text-to-image generation. Generative blocks world~\cite{gbw} represent scene elements as 3D convex primitives and enable control during generation by geometrically manipulating primitives. Both SeeThrough3D and Generative blocks world are limited to controlling layouts of generated images and do not work with real images.

\paragraph{Image Editing with Diffusion Models.}
Recent large text-to-image diffusion models have enabled high-quality image editing~\cite{hertz2022prompt,brooks2023instructpix2pix} thanks to their rich image generation priors. Inference time methods involve inverting real images into the latent space using diffusion inversion~\cite{song2020denoising,kawar2023imagic,mokady2023null}, and then editing is performed by manipulating the cross-attention maps~\cite{hertz2022prompt,masa-ctrl,ye2023ip,avrahami2025stable} during denoising. More recently, training-based methods fine-tune a pretrained diffusion model for instruction-based image editing~\cite{brooks2023instructpix2pix,ominicontrol,ominicontrol2,easycontrol,qwen-image}, scene relighting~\cite{magar2025lightlab}, material control~\cite{sharma2024alchemist,cheng2025marble}, or scene composition~\cite{chen2024anydoor,yang2023paint}. For fine-grained editing, continuous control has been explored to smoothly control individual semantic attributes ~\cite{gandikota2024concept,kamenetsky2025saedit,parihar2025kontinuous}. Another approach aims to personalize the generation given one or a few images of a subject and enable generation in different compositions with text prompts~\cite{galimage,ruiz2023dreambooth,kumari2023multi,kumari2025generating,ye2023ip,garibi2025tokenverse}.

\paragraph{Geometric Editing in Diffusion models.}

Existing methods for geometric image editing rely on text~\cite{spatialedit, obj3dit} or drag based control ~\cite{dragdiffusion,dragondiffusion,gooddrag,clipdrag} or 3D representations of the objects in the scene, either through meshes ~\cite{sam3d,3d-fixer} or built with depth maps and object masks~\cite{diffusion-handles,freefine,geodiffuser,parihar2025zero,blenderfusion,geo-edit}. However, these representations are not user-friendly, require external models~\cite{sam3,depth-anything,unidepth,trellis} and offer limited 3D control. Furthermore, these methods involve a hole-filling procedure via an inpainting model to fill the holes due to the displacement of objects which can result in inconsistent texture and lighting in the scene.  Another line of work leverage supervised learning on video data~\cite{3d-fixup,magic-fixup} to perform image editing, however, these methods often require a chain of intermediate models~\cite{raft,sam3,grounded-sam,instantmesh} to obtain clean supervisory data which is computationally expensive. To overcome all these issues, we design a representation that encodes 3D layouts in a user-friendly manner. Coarse representations such as 3D boxes and convex primitives serve as an effective representation to model object geometry and enable control over 3D scene layout~\cite{seethrough3d,neural-assets,neural-usd,gbw}. Motivated by this, we associate pixels of objects in the scene with an editing-friendly cuboid representation for any scene. 

\begin{figure}[t!]
    \centering
    \includegraphics[width=0.9\linewidth]{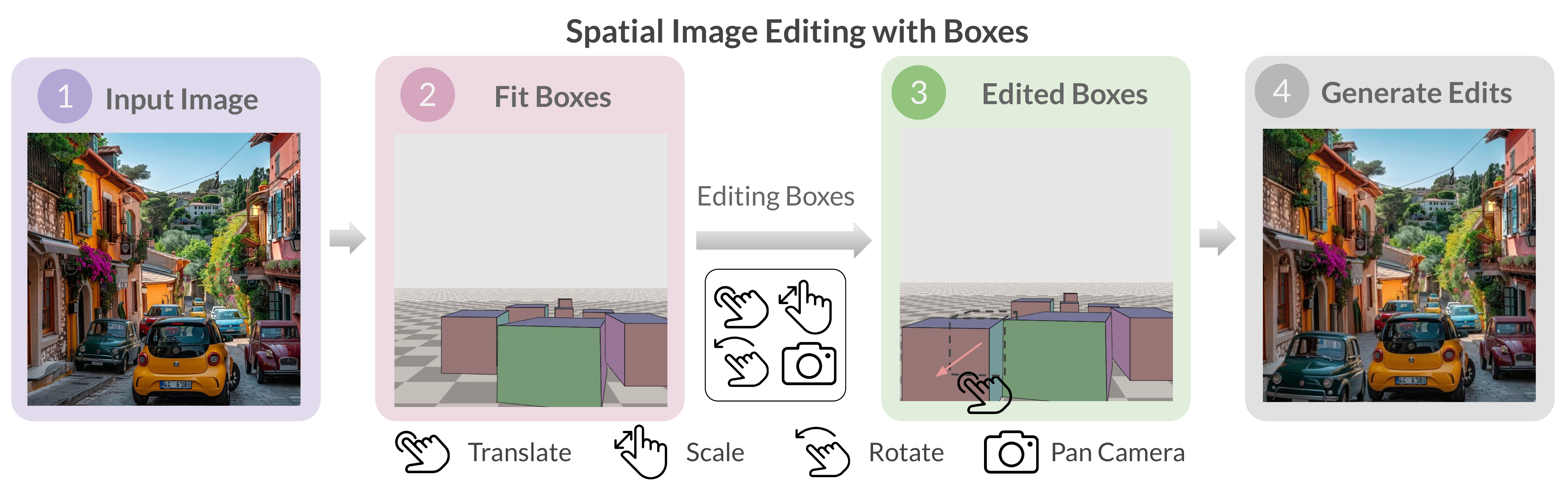}
    \vspace{-10pt}
    \caption{\textbf{Editing Pipeline.} (1) Users provide a real source image and
(2) fit 3D boxes to the objects within the scene using a point-and-click
interface. (3) The boxes can be manipulated in 3D space, allowing for scaling, rotation, translation, and camera moves. Both the source and target box layouts are projected
into 2D and serve, alongside the source image, as inputs to an image
editing model~\cite{flux-kontext}. (4) The model generates an edit
that respects the underlying scene geometry and follows the user's
layout. In this example, the car on the left is
moved forward with the rest of the scene preserved.}
    \label{fig:method-interface}
    \vspace{-10pt}
\end{figure}
\section{Method}
\label{sec:method}

We frame image editing as a transformation between two states of a structured scene representation  (Fig.~\ref{fig:method-interface}). Given a source image $I_{src}$, the user lifts it into a set of 3D boxes, edits the boxes directly, and the resulting box pair conditions an image generator that produces the edited image. 

\paragraph{3D object boxes.}
We represent each object of interest as a 3D box $B_i$ with position, orientation, and scale. To convey 3D orientation through a 2D conditioning signal, we assign a fixed color to each face: red (front), green (back), pink (right), cyan (left), blue (top), and yellow (bottom). The visible face colors in the projected box encode which side of the object faces the camera, independent of object identity or appearance. This color code is canonical across all scenes and is the only mechanism by which the generator reads object orientation.

\paragraph{Scene layout.}
The input to our method -- a set of boxes in two configurations -- has
to tell the generator unambiguously what edit to perform. The mapping
from the source configuration to the target configuration must have a
single interpretation. Boxes alone do not satisfy this: a leftward
shift of every box can equally indicate that the objects moved left
or that the camera panned right, and the generator has no way to
choose. We resolve the ambiguity by anchoring the boxes in a shared
coordinate frame -- a depth-aligned planar floor, rendered as a
checkerboard with depth-aware shading. The floor moves with the
camera and stays fixed under object motion, so any change in the
relative configuration of boxes and floor uniquely identifies what
moved. It also provides a global reference for contact and shadow.
Both boxes and the floor form the \emph{3D scene layout}
$L_{src}$, which we project to 2D as the spatial conditioning image
used by the generator

\paragraph{Fitting boxes to images.}
The user fits boxes to objects through a point-and-click interface (Fig.~\ref{fig:method-interface}). The floor is estimated automatically from the image, so the user only specifies object boxes -- they never author the floor explicitly. To reduce manual effort further, off-the-shelf 3D box detectors~\citep{wilddet3d} produce an initial set of boxes that the user refines.

\paragraph{Editing boxes.}
\begin{wrapfigure}[]{r}{0.5\textwidth}
\vspace{-15pt}
    \centering
    \includegraphics[width=\linewidth]{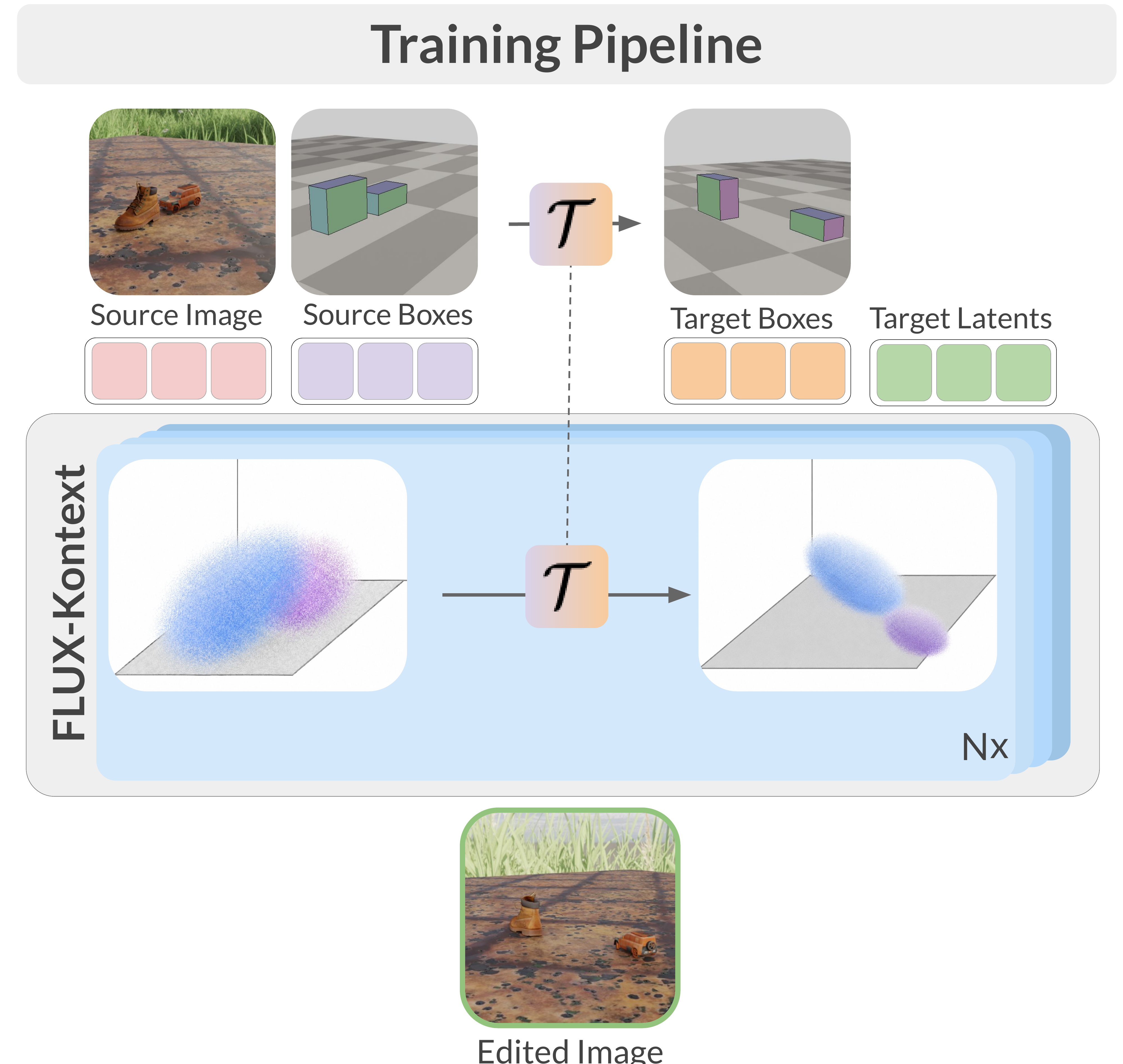}
    \vspace{-15pt}
\caption{\textbf{Training pipeline.} A training example provides a source image,
a noisy target image, and the source and target box projections that
specify the transformation $\mathcal{T}$ between them. All four are
encoded by the VAE and concatenated spatially. Joint attention inside
FLUX-Kontext applies $\mathcal{T}$ to the image latents -- mapping the
source latent to the target latent under box-pair conditioning. LoRA
layers in the attention matrices enable box-conditioned editing while
preserving the base model's prior.}
    \label{fig:method-fig3}
    \vspace{-10pt}
\end{wrapfigure}
The user manipulates the boxes in 3D to specify the desired edit. We support four operations, all expressed as direct manipulation of the same boxes. \emph{Translation} moves a box to a new position while preserving its orientation and scale. \emph{Scaling} resizes a box in place. \emph{Rotation} re-orients a box so that a different colored face is visible to the camera -- e.g., turning a chair from frontal to side view rotates the red face out and the pink or cyan face in. \emph{Camera moves} re-orient all boxes and the floor jointly with respect to a new viewpoint. The result is a target layout $L_{tgt}$, projected to 2D as the target spatial conditioning image. All four edit types are expressed in the same box language; the generator does not need separate handling for translation, rotation, or camera change.

\paragraph{Editing images.}
% \label{sec:method-edit}
We build upon FLUX-Kontext~\citep{flux-kontext, sd3} 
image editor that operates on multimodal token streams. The image $I{src}$, projected source layout $L{src}$ and projected target layout $L_{tgt}$ are first encoded into latent tokens using the VAE and then concatenated along the spatial dimension~\citep{easycontrol,seethrough3d}. All three streams have the same positional encoding, so there is an alignment between image region and the box that covers it. Internally to MMDiT, each stream attends to the others. Image tokens attend to source layout tokens to associate appearances with geometry and to target layout tokens to spatialize appearance in the output. We then decode a single stream of tokens as output representing the edited image.

\paragraph{Training.}
\label{sec:method-train}

We fine-tune FLUX-Kontext with LoRA~\citep{lora} layers injected into the attention matrices, leaving the rest of the model frozen. This adapts the base editor to consume the new box-conditioning streams while preserving its strong generative prior. 

\vspace{-10pt}
% \section{Synthetic dataset}
\section{Dataset}
\label{sec:dataset}
\vspace{-10pt}

For our training, we first create a synthetic dataset of paired views, with same objects existing in two separate 3D positions, and a pair of views that share these objects. Each pair of views is completely calibrated; the transformation from one view's boxes to the other's is provided as supervision.

\paragraph{Scene construction.}
Each scene contains two objects placed on a planar floor under a sampled HDRI. Objects are drawn from the Objaverse~\cite{objaverse-xl} pool, normalized to a common size, and randomly oriented on the ground under a non-overlap constraint. We render the scene from two camera viewpoints. Between the two views, the objects undergo a controlled perturbation -- a fresh rotation, a uniform rescaling, and a translation -- so that the second view contains the same objects in a transformed configuration. The pair therefore supervises both \emph{what edit was made} (the box-pair transformation) and \emph{what the resulting image should look like} (the second view). Two objects per scene is the smallest setting that exercises inter-object occlusion and contextual edits without overwhelming training cost.

\paragraph{Renders and metadata.}
For each view we render a photorealistic RGB image at $512 \times 512$ and a corresponding oriented-box overlay, where the visible faces of each object's 3D box are color-coded by their canonical orientation. These two form the input--conditioning pair used during training.

\paragraph{Scale.}
The dataset comprises {110,000} scenes ({220,000} views) drawn from {10,143} unique 3D objects, paired with {1,154} HDRIs and {5,760} floor materials. 
Additional {10,000} image pairs are taken from the {Objectron}~\cite{objectron} dataset for finetuning. Additional details are in the appendix.
\vspace{-5pt}

\begin{figure}[t]
    \centering
    \includegraphics[width=\linewidth]{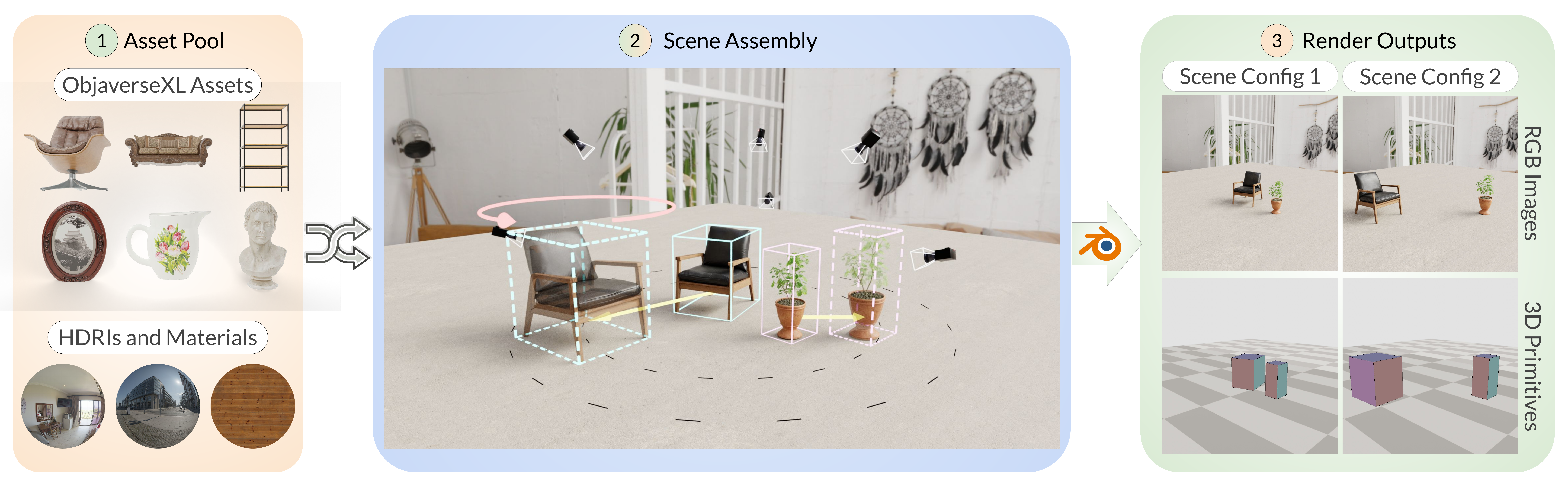}
    \vspace{-15pt}
\caption{\textbf{Dataset rendering pipeline.} \textbf{(1)} 3D assets are
drawn from a subset of Objaverse-XL~\cite{objaverse-xl}; HDRIs and floor
textures are taken from 3D-Fixer~\cite{3d-fixer}. \textbf{(2)} Each
scene places [N] scale-normalized objects on a textured floor under a
sampled HDRI, with collision-aware placement. We then perturb the
objects in rotation $[0, 2\pi]$, scale $[0.5, 1.5]$, and position to
produce a second scene configuration. \textbf{(3)} Blender Cycles
renders both configurations, producing paired RGB images and
color-coded 3D box visualizations along with camera intrinsics,
extrinsics, and per-object material parameters.}
    \label{fig:rendering_pipeline}
\vspace{-15pt}
\end{figure}

\section{Experiments}
\label{sec:experiments}
\vspace{-5pt}
\paragraph{Implementation details:} We use Flux-Kontext~\cite{flux-kontext} as our base image editing model. We initially train for 50$\textit{K}$ steps on a synthetic rendered dataset and then train for another 10$\textit{K}$ steps on a hybrid dataset of 10$\textit{K}$ real samples obtained from objectron and 10$\textit{K}$ synthetic set. 
Additional implementation details can be found in the appendix.

\begin{figure} [h]
    \centering
    \includegraphics[width=\linewidth]{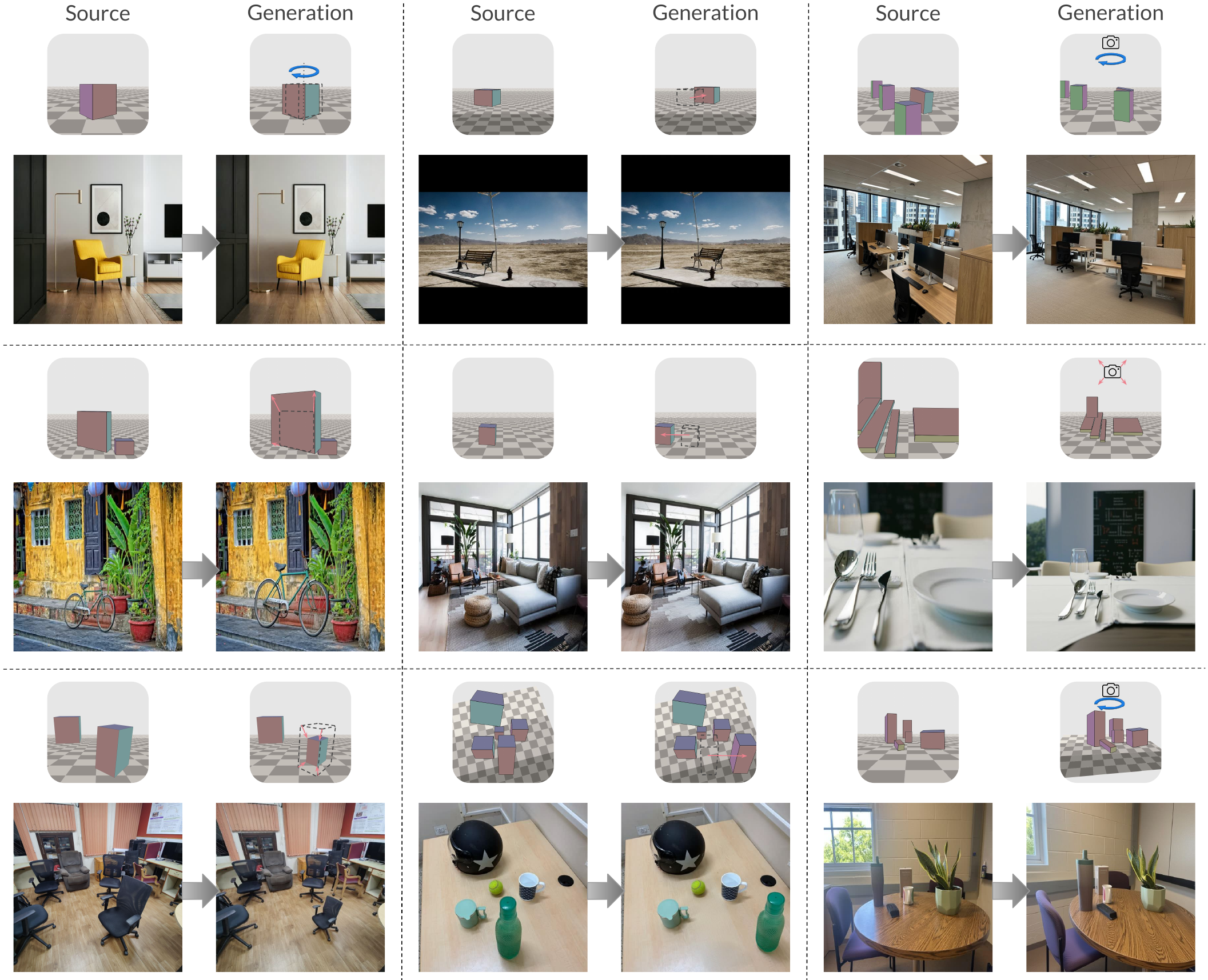}
\vspace{-15pt}
    \caption{\textbf{In-the-wild results.} Nine real-image edits across
diverse indoor and outdoor scenes (offices, dining rooms, street
scenes, desert, living rooms). Each panel shows the source image with
its fitted boxes (top-left inset), the user's edited boxes (top-right
inset), and the resulting generation. Edits include translation,
rotation, scaling, and camera viewpoint changes; the same model
handles all four operations. Scene context, lighting, and occluded
regions are preserved consistently across edits, with no per-image
optimization or fine-tuning. Notice the back of the chair in bottom right figure.}
    \label{fig:quals-fig.8}
\vspace{-15pt}
\end{figure}

\paragraph{Evaluation Dataset}
We evaluate on two settings: \emph{object editing} (objects are
moved while camera is fixed) and \emph{camera editing} (camera moves while objects stay fixed). Each setting is tested on
both synthetic and real data.

\emph{Synthetic.} Both settings use synthetic test sets rendered
with the same pipeline as our training data (Sec.~\ref{sec:dataset}),
held out from training. For object-editing pairs, we apply random translations and rotations to objects; for camera-editing pairs, we apply random viewpoint changes with objects fixed.

\emph{Real.} For object editing, we use
{WildDet-3D}~\cite{wilddet3d}, which provides 3D bounding box
annotations on in-the-wild images. We construct test pairs by
applying random translation and rotation perturbations to each labeled
box, constraining the perturbed box to remain within the image frame.
For camera editing, we use {Objectron}~\cite{objectron} samples held out from training, whose
video sequences naturally provide viewpoint changes; we sample frame
pairs from each video and use the provided per-frame 3D bounding boxes
to construct the source and target layouts. 

\paragraph{Baselines:} For object manipulation (Fig~\ref{fig:object_compare}), we compare against SAM-3D~\cite{sam3d},
3D-Fixer~\cite{3d-fixer}, and SpatialEdit~\cite{spatialedit}, along
with three training-free baselines:
GeoDiffuser~\cite{geodiffuser}, Diffusion Handles~\cite{diffusion-handles},
and FreeFine~\cite{freefine}. For camera editing, we compare
against SpatialEdit~\cite{spatialedit}, SEVA~\cite{seva}, and
Qwen-Camera-Control~\cite{qwen-image}. We additionally compare
qualitatively against GBW~\cite{gbw} in Fig~\ref{fig:gbw-vs-ours}, which operates only on
generated images, on the failure cases reported in their paper.

\vspace{-5pt}
\subsection{Qualitative Results}

We present qualitative results on both real and in-the-wild scenes in Fig.~\ref{fig:quals-fig.8} and Fig.~\ref{fig:quals-fig.7}. 
Our method closely adheres to the targeted layout changes and produces realistic edits across a wide range of inputs. 
Despite being trained predominantly on synthetic data rendered in Blender 
and fine-tuned on only a small amount of real-world data, 
our model generalizes effectively to real scenarios, preserving overall scene integrity 
as well as the fine-grained textures and details of both the edited object and its surroundings. 
As shown in row 3 of Fig.~\ref{fig:quals-fig.7}, the model maintains coherent scene structure even when multiple edits are applied simultaneously. 
Row 1 of the same figure further demonstrates that the model accurately follows edit instructions on a sketch-style input, despite this domain being entirely out of distribution.

\vspace{-5pt}
\subsection{User Study}

\begin{wrapfigure}[10]{r}{0.5\textwidth}
% \begin{figure}
\vspace{-30pt}
    \centering
    \includegraphics[width=\linewidth]{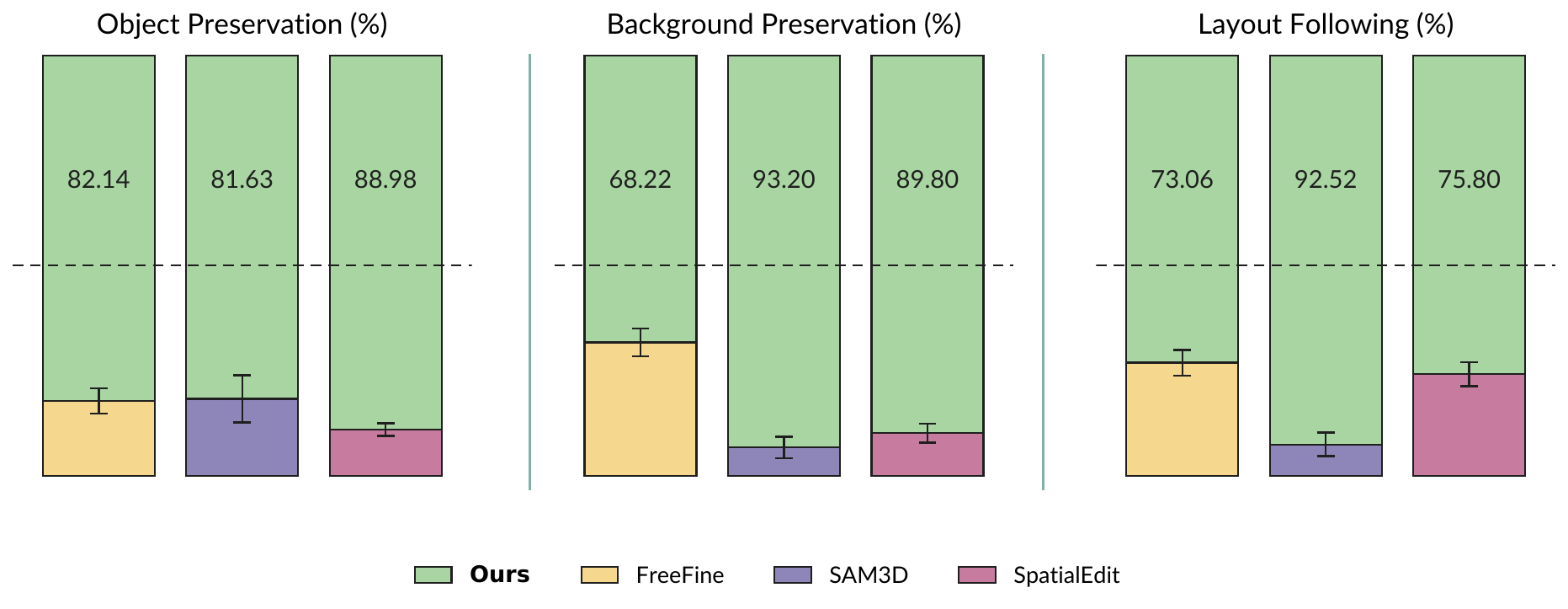}
    \vspace{-20pt}
    \caption{\textbf{User Study.} Each bar represents user preference rate for our method compared to baselines, categorized by specific evaluation criteria.}
    \label{fig:user_study}
    \vspace{-20pt}
% \end{figure}
\end{wrapfigure}

We conducted an A/B user study with 49 participants, each asked to select 
the preferred output from randomly sampled generations produced by our 
method and three baselines: FreeFine~\cite{freefine}, SAM3D~\cite{sam3d}, 
and SpatialEdit~\cite{spatialedit}. We evaluate a) object preservation, b) background preservation, and c) layout 
following. Results highlight high preference rates for our method in all evaluation categories (see Fig.\ref{fig:user_study})

\begin{figure}[p]
% \vspace{-5pt}
    \centering
    \includegraphics[width=0.99\linewidth]{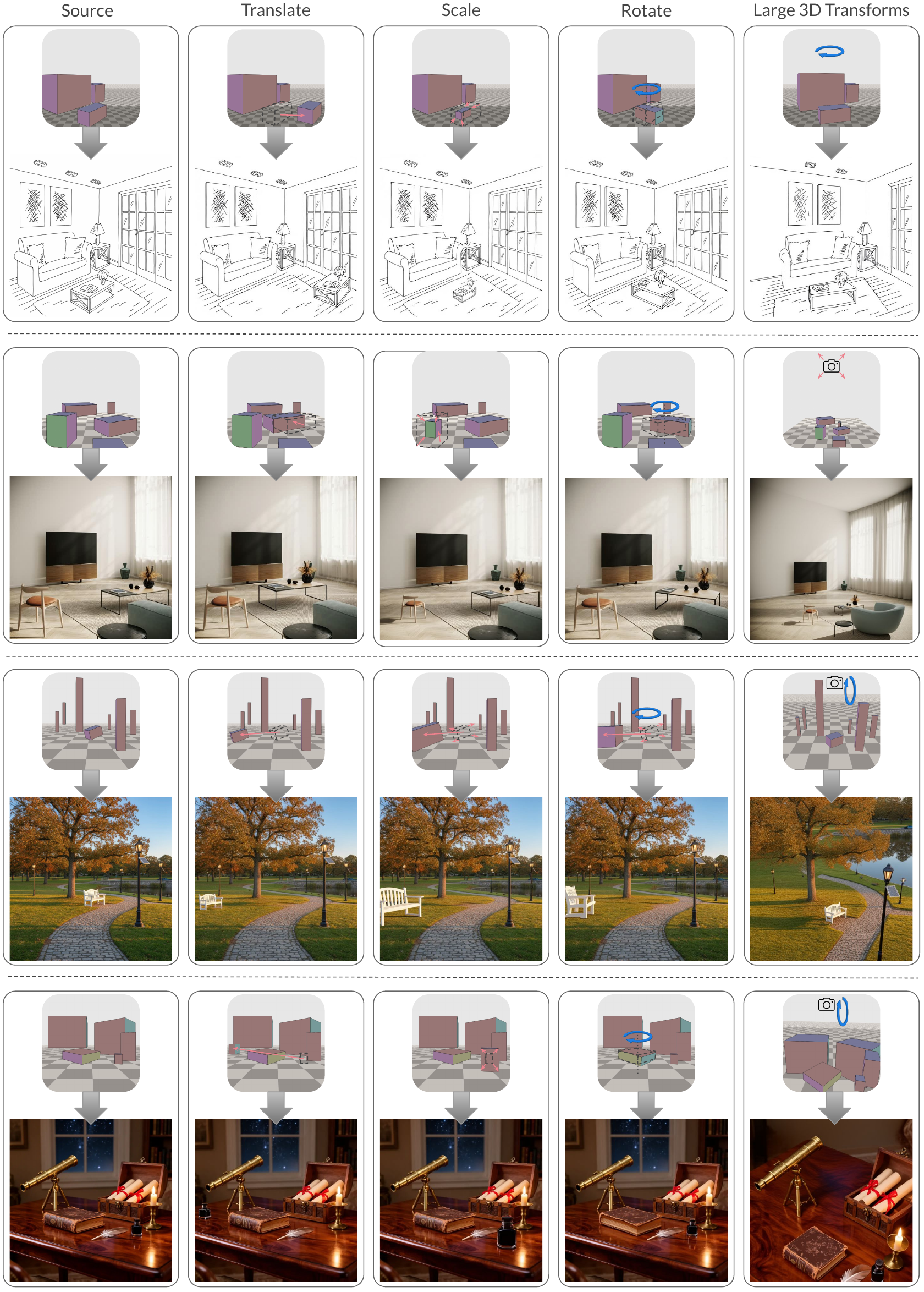}
% \vspace{-10pt}
\caption{\textbf{Scene editing across operations, scenes, and rendering
styles.} Each row shows a source image (column 1) edited under four
operations: translation, scaling, rotation, and a large combined
transformation (columns 2--5). The same trained model handles all
operations across all scenes -- including the line drawing in row 1,
despite training only on photographic data. The model preserves
rendering style under the edit (row 1 stays a line drawing) and
respects scene-level effects that prior methods typically destroy:
shadow and lighting consistency under viewpoint change in the autumn
park (row 3), and fire and metallic reflections on polished surfaces
in the table with book scenes (row 4). Best viewed on screen and zoom in to see all edits clearly.}
    \label{fig:quals-fig.7}
    % \vspace{-20pt}
\end{figure}

\vspace{-5pt}
\subsection{Metrics}
\label{sec:metrics}
Following FreeFine~\cite{freefine}, we evaluate each baseline's edited output $I_e$ against the ground-truth target $I_{gt}$ using eight metrics that span image quality, region-localized consistency, and edit fidelity. For image quality, we report PSNR\,$\uparrow$, SSIM~\cite{SSIM}\,$\uparrow$, LPIPS~\cite{LPIPS}\,$\downarrow$, and DreamSim~\cite{dreamsim}\,$\downarrow$. For region-localized consistency, we measure subject and background consistency as masked DINOv3-ViT-B/16~\cite{dinov3} feature similarity computed under the foreground mask and its complement, respectively. For edit fidelity, we report four metrics: \textbf{Warp Error}, the masked $L_1$ difference between $I_e$ and a reference target; \textbf{Mean Distance}, the pixel error between DIFT~\cite{dift} semantic correspondences computed from source-to-$I_e$ and source-to-reference; \textbf{IoU}, the intersection-over-union between the SAM3~\cite{sam3} mask of the generated object and the target bounding box; and \textbf{Angular Error}, between target and generated orientations.

\begin{figure}[t!]
    \centering
    \includegraphics[width=\linewidth]{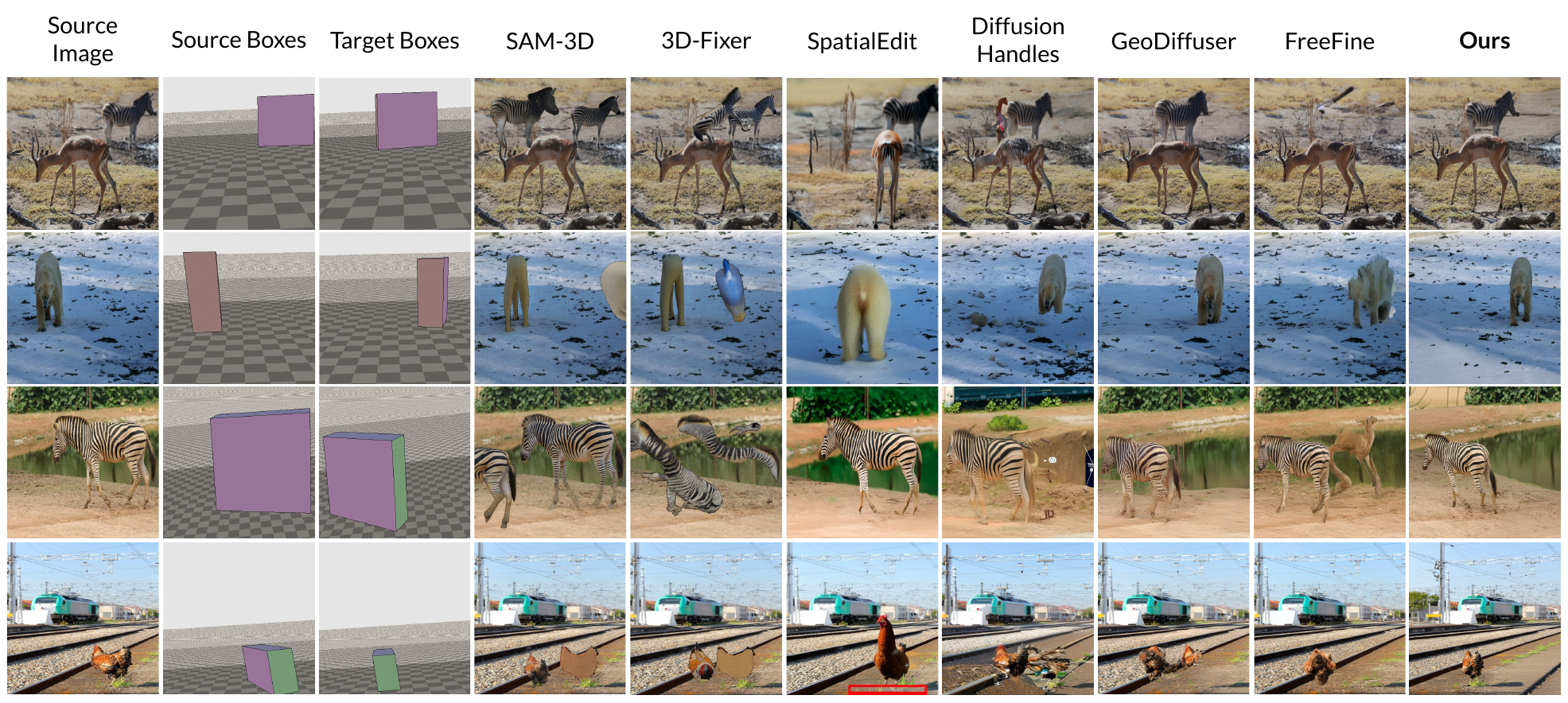}
    \vspace{-15pt}
\caption{\textbf{Qualitative comparison on real images.} For each
example, the user specifies an edit by editing 3D boxes (columns
2--3), which we render to 2D as the target conditioning input for our
method and convert to the appropriate input format for each baseline
(mask, depth, control points). The examples shown are deliberately
hard: every subject is a non-rigid animal (zebras, impala, polar bear,
chicken), and the edits include rotations and
disocclusions. These cases break the assumptions of methods that fit
3D meshes or warp pixels via depth, since deformable objects do not
admit clean rigid transformations in image space. Baselines~\cite{
sam3d,3d-fixer,spatialedit,diffusion-handles,geodiffuser,freefine}
accordingly fail to apply the edit, introduce identity drift, or
destroy scene context. Our method (rightmost column) handles the same
edits cleanly, recovering object regions not visible in the input without any hallucinations.}
\vspace{-15pt}
\label{fig:object_compare}
\end{figure}

\vspace{-5pt}
\subsection{Object Control in Real-World Scenes}
% \vspace{-5pt}

\begin{wrapfigure}[]{r}{0.3\textwidth}
\vspace{-15pt}
% \begin{figure}
    \centering
    \includegraphics[width=\linewidth]{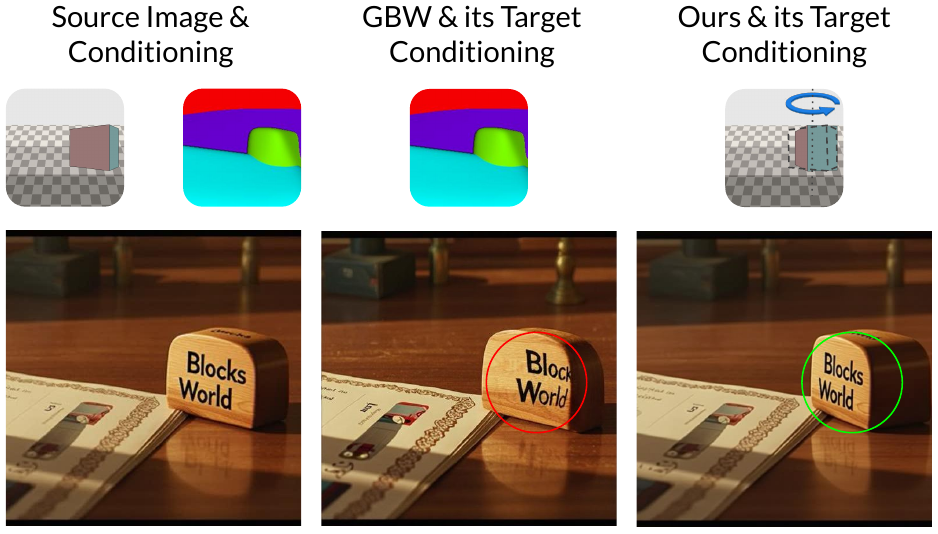}
\vspace{-15pt}
 \caption{\small \textbf{GBW's reported failure cases.} This is an example GBW~\cite{gbw} flags as difficult for their method. Our box-space conditioning handles it, while their depth-and-warp convex primitive pipeline does not. Source (left), GBW (middle), ours (right).
 }
    \label{fig:gbw-vs-ours}
\end{wrapfigure}

We report quantitative metrics against both 3D editing and image editing baselines. Results are shown in Tab.~\ref{tab:real_edit_object} and Tab.~\ref{tab:syn_edit_object}.
Our method significantly outperforms prior work, including SAM3D~\cite{sam3d}, 3D-Fixer~\cite{3d-fixer}, SpatialEdit~\cite{spatialedit}, DiffusionHandles~\cite{diffusion-handles}, GeoDiffuser~\cite{geodiffuser}, and FreeFine~\cite{freefine}. The edits are temporally consistent across views and geometrically faithful to the target transformation. Our method also respects the intended placement of 3D objects in space.
On real object editing (Tab.~\ref{tab:real_edit_object}), our method ranks first or second on every metric. When second, it remains close to the top baseline. When first, it leads by a clear margin particularly in mean distance error and angular error.
On synthetic object editing (Tab.~\ref{tab:syn_edit_object}), our method outperforms all baselines across every metric. Qualitative comparisons are provided in Fig.~\ref{fig:object_compare}.
Additional results are discussed in the appendix.

\begin{table*}[h!]
\vspace{-10pt}
\centering
\caption{Quantitative comparison on \emph{object editing} using WildDet-3D~\cite{wilddet3d} data.}
\label{tab:real_edit_object}
\begin{adjustbox}{max width=\textwidth}
\begin{tabular}{l c c c c c c}
\toprule
\multirow{2}{*}{Method}
& \multicolumn{2}{c}{Consistency (DINO)}
& \multicolumn{4}{c}{Edit fidelity} \\
\cmidrule(lr){2-3}\cmidrule(lr){4-7}
& Subject \up & Background \up
& Warp Error \dn & Mean Distance \dn & IoU \up & Angular Error \dn \\
\midrule
SAM3D~\cite{sam3d}                        & 0.632 & 0.956 & 0.175 & 79.087 & 0.356 & 99.239 \\
3D-Fixer~\cite{3d-fixer}                  & 0.570 & 0.942 & 0.165 & 74.294 & 0.361 & 108.762 \\
SpatialEdit~\cite{spatialedit}            & 0.581 & 0.884 & 0.234 & 107.438 & \textbf{0.399} & 98.171 \\
DiffusionHandles~\cite{diffusion-handles} & 0.631 & 0.924 & 0.154 & \underline{32.634} & 0.369 & 95.659 \\
GeoDiffuser~\cite{geodiffuser}            & \textbf{0.646} & 0.953 & 0.143 & 35.488 & 0.373 & \underline{94.824} \\
FreeFine~\cite{freefine}                  & 0.623 & \textbf{0.956} & \underline{0.143} & 42.125 & 0.382 & 98.705 \\
\midrule
Ours                                      & \underline{0.637} & \underline{0.955} & \textbf{0.136} & \textbf{25.910} & \underline{0.382} & \textbf{92.051} \\
\bottomrule
\end{tabular}
\end{adjustbox}
% \vspace{-10pt}
\end{table*}

\begin{table*}[h!]
\vspace{-10pt}
\centering
\caption{Quantitative comparison on \emph{object editing} using synthetic test set.}
\label{tab:syn_edit_object}
\begin{adjustbox}{max width=\textwidth}
\begin{tabular}{l c c c c c c c c c c}
\toprule
\multirow{2}{*}{Method}
& \multicolumn{4}{c}{Image quality}
& \multicolumn{2}{c}{Consistency (DINO)}
& \multicolumn{4}{c}{Edit fidelity} \\
\cmidrule(lr){2-5}\cmidrule(lr){6-7}\cmidrule(lr){8-11}
& PSNR \up & SSIM \up & LPIPS \dn & DreamSim \dn
& Subject \up & Background \up
& Warp Error \dn & Mean Distance \dn & IoU \up & Angular Error \dn \\
\midrule
SAM3D~\cite{sam3d}                        & 20.415 & 0.810 & 0.208 & \underline{0.195} & \underline{0.769} & \underline{0.931} & 0.173 & \underline{44.557} & \underline{0.374} & 123.838 \\
3D-Fixer~\cite{3d-fixer}                  & 20.822 & \underline{0.812} & \underline{0.203} & 0.207 & 0.724 & 0.929 & 0.170 & 47.110 & 0.368 & 126.193 \\
SpatialEdit~\cite{spatialedit}            & 12.194 & 0.433 & 0.601 & 0.391 & 0.608 & 0.792 & 0.262 & 100.287 & 0.274 & 128.287 \\
DiffusionHandles~\cite{diffusion-handles} & 18.717 & 0.693 & 0.366 & 0.347 & 0.651 & 0.828 & 0.146 & 55.535 & 0.357 & 125.289 \\
GeoDiffuser~\cite{geodiffuser}            & 20.654 & 0.740 & 0.275 & 0.282 & 0.668 & 0.876 & 0.146 & 57.451 & 0.365 & 124.215 \\
FreeFine~\cite{freefine}                  & \underline{20.852} & 0.777 & 0.226 & 0.222 & 0.698 & 0.905 & \underline{0.145} & 55.997 & 0.359 & \underline{122.606} \\
\midrule
Ours                                      & \textbf{23.686} & \textbf{0.821} & \textbf{0.130} & \textbf{0.092} & \textbf{0.879} & \textbf{0.964} & \textbf{0.101} & \textbf{21.392} & \textbf{0.534} & \textbf{76.739} \\
\bottomrule
\end{tabular}
\end{adjustbox}
\vspace{-10pt}
\end{table*}

\vspace{-5pt}
\subsection{Ablations}
We ablate design choices of our box conditioning: (1) the \textbf{full setup (ours)}, in which objects are placed on a checkered-pattern floor and the conditioning boxes are rendered from different viewpoints with directional color-coded faces; (2) \textbf{no floor}, where the checkered floor is removed and only the boxes are rendered, leaving the model without an explicit spatial reference for the scene layout; and (3) \textbf{uniform box color}, where all faces of the conditioning box share a single color, removing the directional cue that encodes object orientation.

All three variants are trained on the same 10$\textit{K}$ pairs with identical object placements and lighting; only the floor and box appearance differ. Each model is evaluated on 500 test samples, with quantitative results reported in Table~\ref{tab:ablation_real_camera}

\begin{wraptable}{r}{0.5\textwidth} % Adjust 'r' and '0.5\textwidth' as needed
\vspace{-25pt}
    \centering
    \caption{Ablation on real-set camera edits.}
    \label{tab:ablation_real_camera}
    \resizebox{\linewidth}{!}{% % Use \linewidth to fit the wrap container
        \begin{tabular}{lcccccc}
            \toprule
            Variant & Subj.$\uparrow$ & Bg.$\uparrow$ & Warp$\downarrow$ & MeanDist$\downarrow$ & IoU$\uparrow$ & Angular.$\downarrow$ \\
            \midrule
            nofloor   & 0.562 & 0.829 & 0.174 & 47.667 & 0.381 & \textbf{102.667} \\
            samecolor & \underline{0.568} & \textbf{0.882} & \underline{0.169} & 57.618 & \textbf{0.399} & 105.472 \\
            original  & \textbf{0.574} & \underline{0.870} & \textbf{0.153} & \textbf{42.616} & \underline{0.389} & \underline{103.771} \\
            \bottomrule
        \end{tabular}%
    }
    \vspace{-15pt}
\end{wraptable}

\paragraph{Effect of the checkered floor.}
Without the checkered floor, the model loses a key cue for localizing objects relative to their surroundings, leading to noticeably worse position preservation. We attribute this to the floor pattern providing a consistent global reference frame that anchors the model's sense of spatial layout.

\paragraph{Effect of directional box coloring.}
Replacing the color-coded box faces with a uniform color degrades orientation accuracy: the model can still localize the object, but tends to misread its facing direction. This confirms that the per-face coloring is the primary signal through which orientation information is injected.

Visualizations demonstrating effectiveness of providing both the checkered floor and the direction specific box colors are shown in the appendix. Together, these ablations show that the floor pattern and directional box coloring contribute complementary signals, the global position and local orientation respectively, and that both are necessary to achieve the full performance of our method.
\section{Conclusion and Limitations}
\label{sec:conclusion}

\vspace{-5pt}
We propose an effective and intuitive primitive based image editing framework that enables 3D aware scene edits such as object translation, rotation and camera control. Given an input image, we fit 3D box primitives on individual objects and place the primitives on a floor with checkered pattern. This primitive based representation can be edited easily by dragging or scaling the primitive boxes. Training an image editing model on the source image, its primitives and target primitives results in precise spatial control of the scene objects. To train this model, we generate a synthetic dataset where 3D assets are placed in a rendering engine and captured from different viewpoints in various configurations. Though trained on synthetic dataset, our method works exceptionally well for real world image editing and performs accurate 3D aware edits. This suggests that the diffusion editing models inherently have object aware representations that can be exposed as a simple user interface with small fine-tuning. This opens up new direction for understanding representations inside the foundation image editing models and how to expose them for downstream tasks.

\begin{figure}[H]
    \centering
    \includegraphics[width=0.5\linewidth]{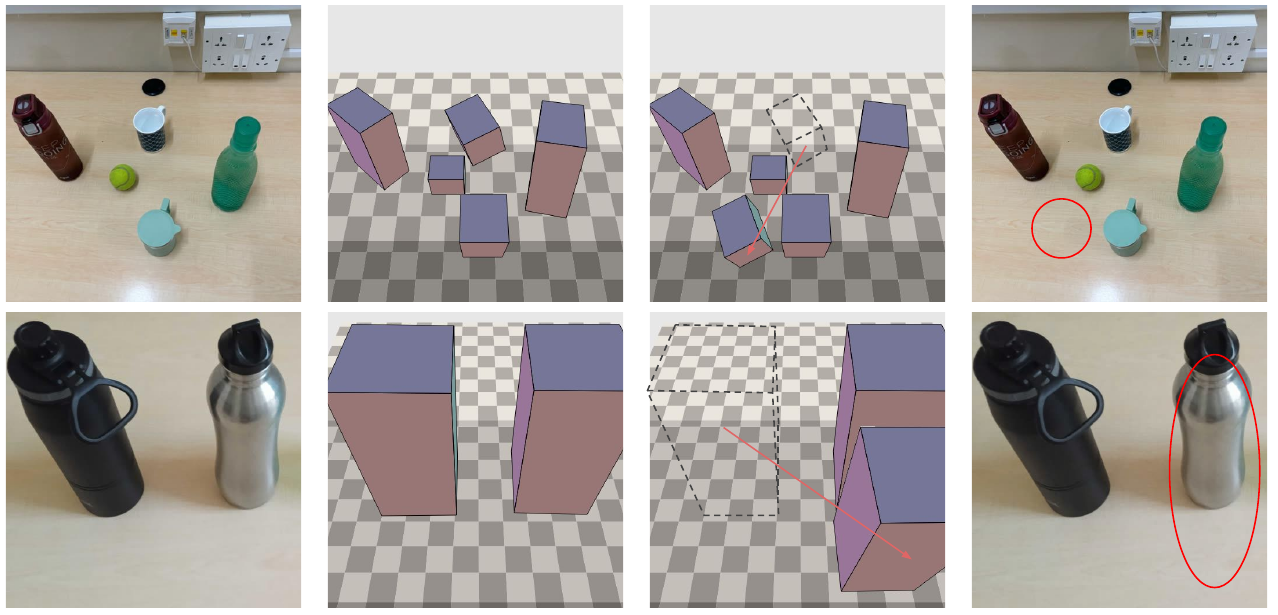}
    \vspace{-10pt}
    \caption{\textbf{Limitation:} Our model fails when multiple objects could plausibly satisfy the target layout. In both examples, either object could be moved to match the target boxes (small objects, top row; large objects, bottom row), and the model refuses to commit to one -- producing an identity transformation instead.}
    \label{fig:limitation}
% \vspace{-10pt}
\end{figure}

\paragraph{Limitation.}  When objects share similar scales, their bounding boxes become indistinguishable. This creates prompt ambiguity, as the final state could represent several different transformations. Consequently, the model fails in these cases, producing only the identity transformation (see Fig.~\ref{fig:limitation}). Moreover, a common limitation of current 3D-editing frameworks is the inability to perfectly isolate modifications. The editing process frequently introduces unintended artifacts and minor pixel-level alterations to the background, failing to maintain strict background consistency. Addressing these unintended background alterations is a highly valuable direction for future research.

% \begin{wrapfigure}[]{r}{0.5\textwidth}
% \vspace{-10pt}
% \end{wrapfigure}

\paragraph{Acknowledgement.}  This work is supported by PMRF by Govt. of India.

\bibliographystyle{plain} % Or 'unsrtnat' for citation order
\bibliography{main}

%%%%%%%%%%%%%%%%%%%%%%%%%%%%%%%%%%%%%%%%%%%%%%%%%%%%%%%%%%%%

\newpage
\appendix
\section*{\centering{Supplemental Materials}}
\vspace{-2mm}
\section*{Table of Contents}
% \vspace{-1em}
\begin{enumerate}[label=\Alph*.]
   \item \hyperlink{sec:supp:addn-results}{\textbf{Additional Results}} \dotfill \pageref{sec:supp:addn-results}
    \item \hyperlink{sec:supp-dataset}{\textbf{Dataset Generation: Implementation Details}} \dotfill \pageref{sec:supp:dataset}
    \item \hyperlink{sec:supp:impln}{\textbf{Implementation Details}} \dotfill \pageref{sec:supp:impln}
    \item \hyperlink{sec:supp:ablations}{\textbf{Ablation Visualizations}} \dotfill \pageref{sec:supp:ablations}
    \item \hyperlink{sec:supp:user-study}{\textbf{User Study Setup}} \dotfill \pageref{sec:supp:user-study}
\end{enumerate}

\section{Additional results}
\label{sec:supp:addn-results}

\subsection{Qualitative results}

\begin{figure}[h!]
    \centering
    \includegraphics[width=\linewidth]{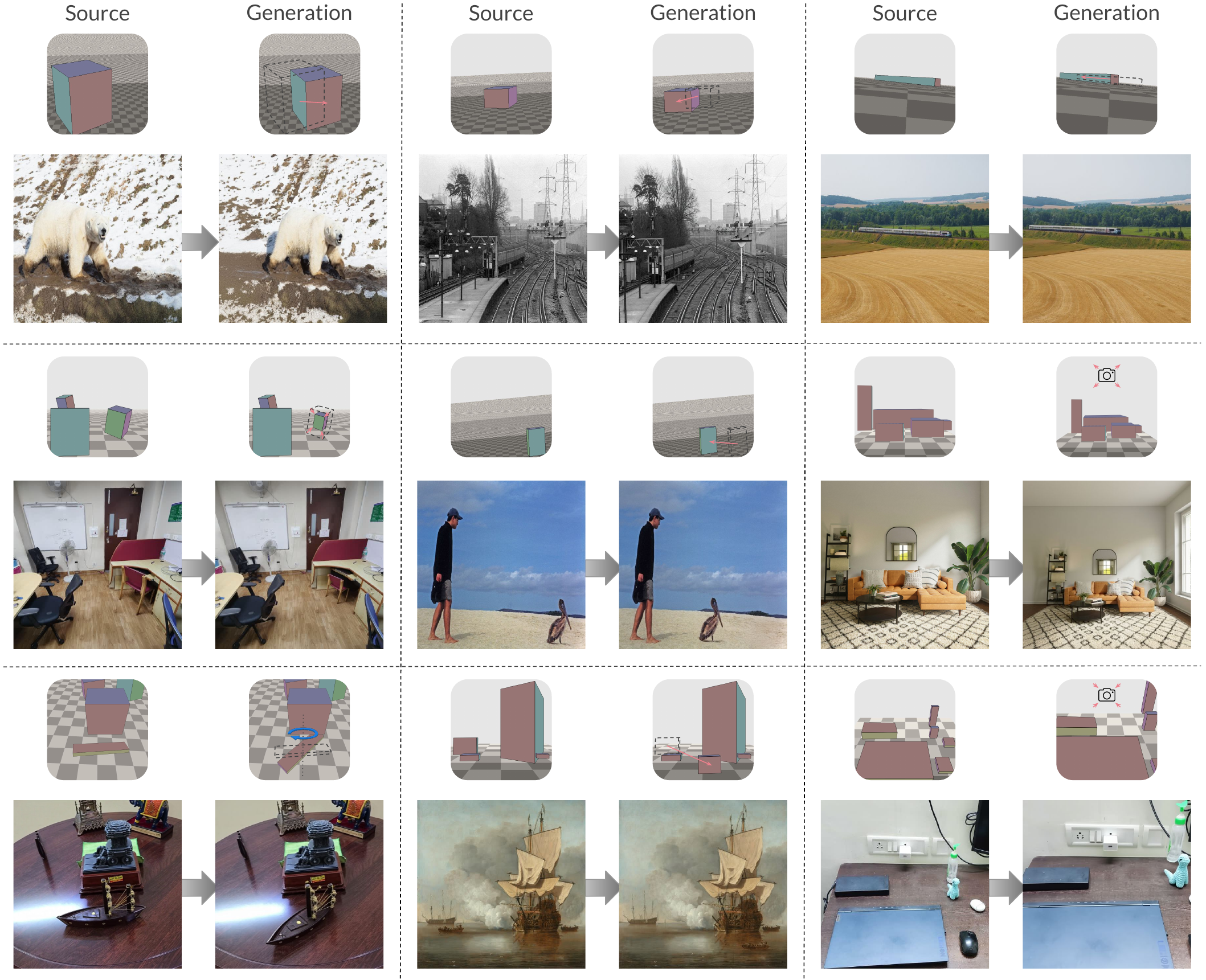}
\vspace{-15pt}
    \caption{\textbf{In-the-wild 3D editing:} We demonstrate the robustness of our method by showcasing additional 3D edits across real world scenes having different lighting, viewpoints and styles. Our method although trained on one or two object scenes, generalizes to complex multi object layouts in real images. Notably we demonstrate all 3D edit operations in this figure from object-centric translation, scaling and rotation to changing camera viewpoints with a single checkpointing, demonstrating the strength of our box conditioning.}
    \label{fig:suppl-quals-fig.1}
\end{figure}

We provide further results in Fig.\ref{fig:suppl-quals-fig.1} and \ref{fig:suppl-quals-fig.2}, showcasing robust 3D scene editing across multiple styles, perspectives and operations. 

\begin{figure}[h]
    \centering
    \includegraphics[width=1\linewidth]{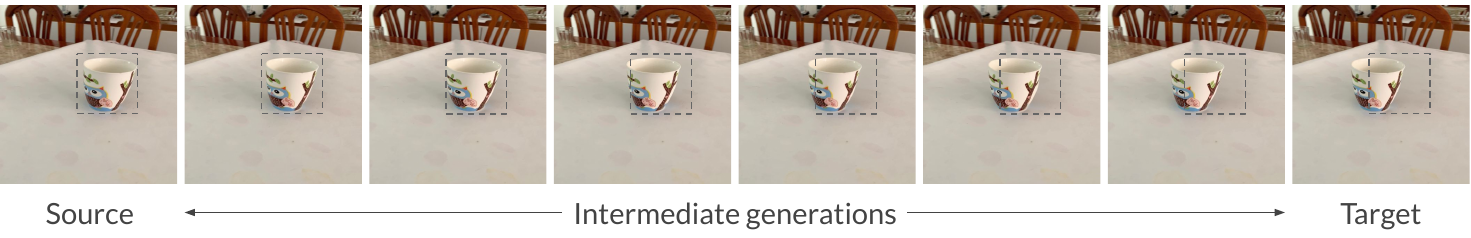}
\vspace{-15pt}
    \caption{\textbf{Smooth Interpolation}: We demonstrate smooth interpolation between the source image and target layout. We overlay the source object position in terms of 2D bounding box on each image to clearly distinguish the incremental layout adjustments and this suggests that the model has internally learnt how to adjusts objects smoothly through box transformations.}
    \label{fig:suppl-interp}
\end{figure}

\begin{figure}[p]
    \centering
    \includegraphics[width=\linewidth]{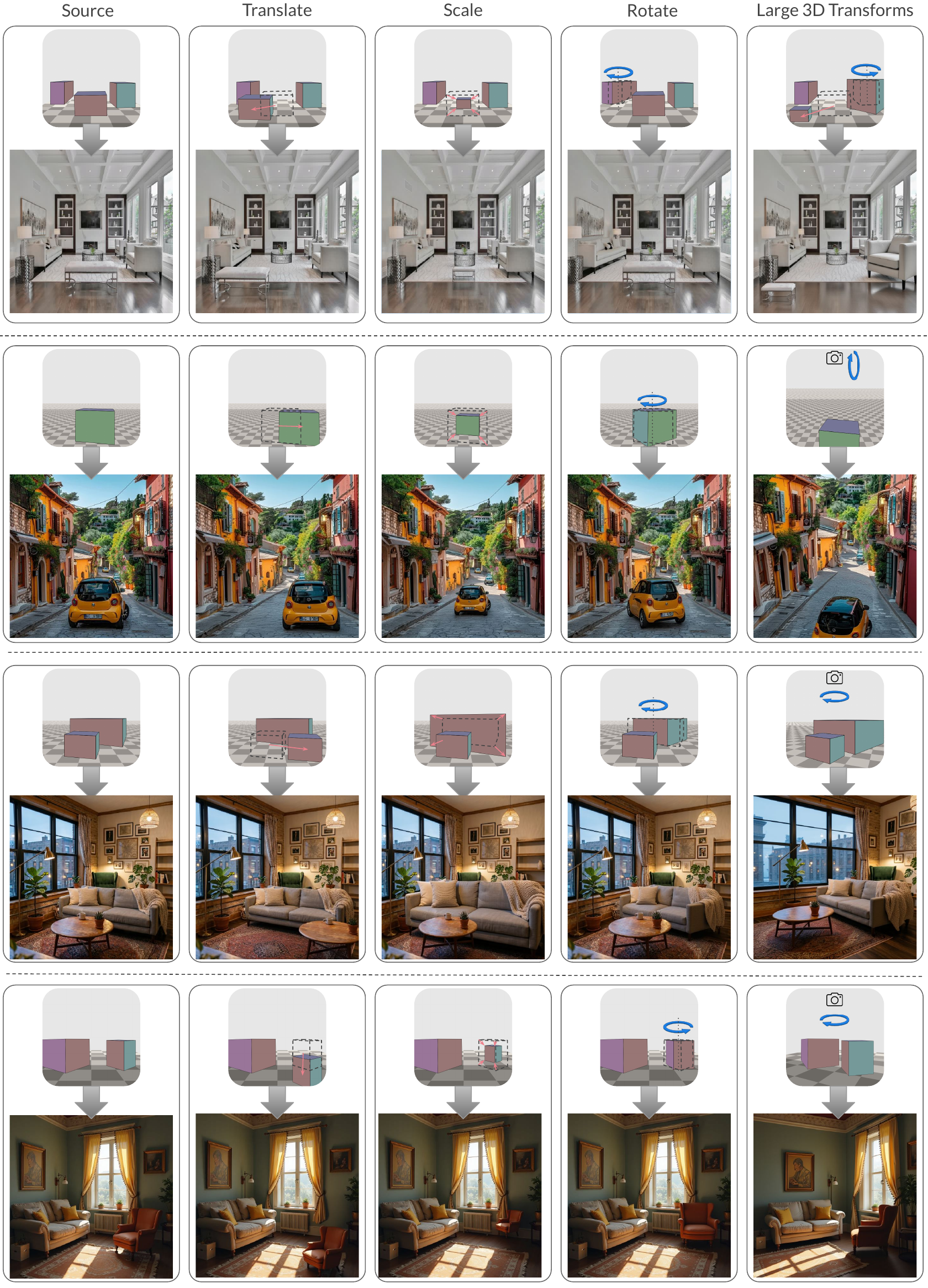}
\vspace{-15pt}
    \caption{\textbf{Rethinking a scene:} This figure illustrates our model’s ability to manipulate 3D scene layouts using primitive box representations. We demonstrate precise control over object location, orientation, and scale, enabling complex 3D transformations such as extreme viewing angles that require the seamless reorganization of multiple scene elements. Ultimately, our method serves as a robust tool for performing diverse, high-fidelity 3D edits on real-world scenes.}
    \label{fig:suppl-quals-fig.2}
\end{figure}

As shown in Fig.\ref{fig:suppl-interp}, our model also maintains high-fidelity details even when applying small, incremental shifts to object positions.

\begin{figure}[H]
    \centering
    \includegraphics[width=0.8\linewidth]{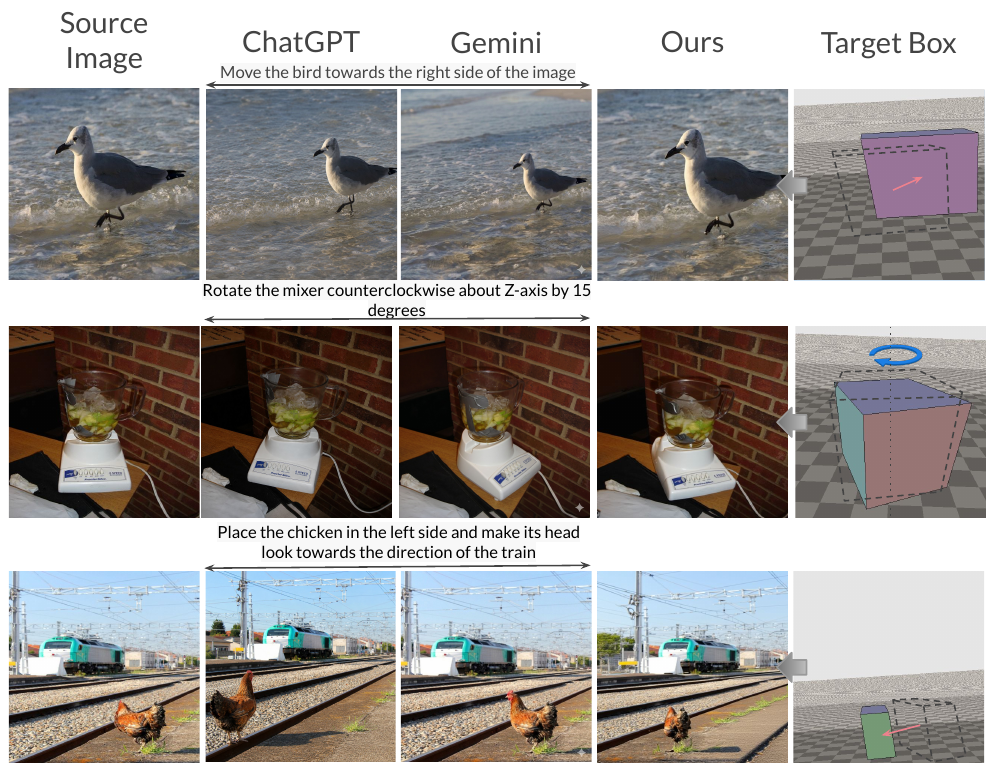}
\vspace{-10pt}
    \caption{\textbf{Comparison with SOTA Models.} We evaluate our method against leading proprietary image generative models, Gemini and ChatGPT, on 3D editing tasks within real-world scenes. Since these models primarily rely on text-based conditioning, which lacks the spatial granularity necessary for 3D manipulation, they often fail to execute the precise 3D edits required by users.}
    \label{fig:suppl-sota}
\end{figure}

Additionally, we compare our model's results with leading proprietary generative models on 3D edits with real-world scenes, As seen in Fig.\ref{fig:suppl-sota}, they struggle to perform precise 3D edits, primarily due to their limitation in terms of text-based conditioning. Our method is able to perform precise 3D edits on these scenes that are otherwise challenging for proprietary models.

\subsection{Quantitative results}

In the camera edit benchmark (Tab.~\ref{tab:real_edit_camera}), our method outperforms all baselines by a significant margin. It consistently preserves scene structure across edits.
On the synthetic camera edit benchmark (Tab.~\ref{tab:syn_edit_camera}), our method again surpasses all baselines. The only exception is background preservation, where we rank a close second to SpatialEdit~\cite{spatialedit}.

\begin{table*}[h]
\centering
\caption{Quantitative comparison on \emph{camera editing} using synthetic test set.}
\label{tab:syn_edit_camera}
\begin{adjustbox}{max width=\textwidth}
\begin{tabular}{l c c c c c c c c c c}
\toprule
\multirow{2}{*}{Method}
& \multicolumn{4}{c}{Reconstruction quality}
& \multicolumn{2}{c}{Consistency (DINO)}
& \multicolumn{4}{c}{Edit fidelity} \\
\cmidrule(lr){2-5}\cmidrule(lr){6-7}\cmidrule(lr){8-11}
& PSNR \up & SSIM \up & LPIPS \dn & DreamSim \dn
& Subject \up & Background \up
& Warp Error \dn & Mean Distance \dn & IoU \up & Angular Error \dn \\
\midrule
SEVA~\cite{seva}                    & \underline{13.726} & \underline{0.460} & \underline{0.554} & \underline{0.397} & 0.571 & 0.783 & 0.210 & \underline{118.374} & 0.207 & 118.039 \\
SpatialEdit~\cite{spatialedit}      & 12.271 & 0.420 & 0.622 & \underline{0.345} & \underline{0.642} & \textbf{0.892} & 0.228 & 123.629 & 0.199 & 118.236 \\
Qwen-Camera-LoRA                    & 13.280 & 0.432 & 0.645 & 0.480 & 0.526 & 0.683 & \underline{0.203} & 137.366 & \underline{0.220} & \underline{117.583} \\
\midrule
Ours                                & \textbf{17.449} & \textbf{0.575} & \textbf{0.351} & \textbf{0.203} & \textbf{0.649} & \underline{0.857} & \textbf{0.130} & \textbf{48.298} & \textbf{0.239} & \textbf{112.358} \\
\bottomrule
\end{tabular}
\end{adjustbox}
\end{table*}

% camera control objectron 
\begin{table*}[h]
\centering
\caption{Quantitative comparison on \emph{camera editing} using \textbf{Objectron}~\cite{objectron} test set.}
\label{tab:real_edit_camera}
\begin{adjustbox}{max width=\textwidth}
\begin{tabular}{l c c c c c c c c c c}
\toprule
\multirow{2}{*}{Method}
& \multicolumn{4}{c}{Reconstruction quality}
& \multicolumn{2}{c}{Consistency (DINO)}
& \multicolumn{4}{c}{Edit fidelity} \\
\cmidrule(lr){2-5}\cmidrule(lr){6-7}\cmidrule(lr){8-11}
& PSNR \up & SSIM \up & LPIPS \dn & DreamSim \dn
& Subject \up & Background \up
& Warp Error \dn & Mean Distance \dn & IoU \up & Angular Error \dn \\
\midrule
SpatialEdit~\cite{spatialedit}              & 8.364 & 0.239 & 0.801 & 0.768 & 0.378 & 0.552 & 0.332 & 116.637 & \underline{0.208} & 126.220 \\
Qwen-Camera-LoRA~\cite{qwen-image}      & 8.854 & 0.293 & 0.807 & 0.778 & 0.372 & 0.532 & 0.307 & 134.394 & 0.166 & 123.095 \\
SEVA~\cite{seva}                            & \underline{10.783} & \underline{0.338} & \underline{0.655} & \underline{0.389} & \underline{0.548} & \underline{0.795} & \underline{0.266} & \underline{104.916} & 0.206 & \underline{118.000} \\
\midrule
Ours                                        & \textbf{15.076} & \textbf{0.452} & \textbf{0.379} & \textbf{0.124} & \textbf{0.791} & \textbf{0.931} & \textbf{0.141} & \textbf{29.406} & \textbf{0.327} & \textbf{70.152} \\
\bottomrule
\end{tabular}
\end{adjustbox}
\end{table*}

\section{Dataset Generation: Implementation Details}
\label{sec:supp:dataset}
 
Generation is split into two decoupled stages. \textbf{Stage~1} (composition) lays objects out and synthesizes camera poses, and this scene configuration is saved; no images are
produced. \textbf{Stage~2} (rendering) reads each scene configuration and produces
all photometric and geometric outputs.
 
\paragraph{Source assets.}
3D models come from Objaverse-XL~\cite{objaverse-xl} and we use a subset of 10\textit{K} assets. We use 1\textit{K} HDRIs and 5.5\textit{K} PBR floor materials following 3D-Fixer~\cite{3d-fixer}.

\paragraph{Stage~1: Composition.}
Each scene contains $N{=}2$ objects on a planar floor, generated inside
BlenderProc~\cite{blenderproc}. Sampled objects are normalized so their
longest side equals $2.5$~units, randomly rotated about the vertical
axis, and placed using an XY circular-bound proxy with a $0.2$ unit
collision margin. Two $C{=}2$ camera poses on a lateral orbit
at distance $d=(r\cdot 1.35)/\tan(\phi/2)$, where $r$ is the joint
bounding-sphere radius and $\phi$ the narrower field of view.To ensure full visibility, all 8 bounding box vertices of every object must project at least 10 pixels inside the frame. The second
view introduces per-object perturbations, including a new azimuth, a uniform rescale between $[0.5,1.5]$, and a small XY translation, while adhering to the original collision and visibility constraints.
 
\paragraph{Stage~2: Rendering.} Fig.\ref{fig:rendering_pipeline} showcases our rendering pipeline.
Scenes are rendered with Cycles on OPTIX at
$512{\times}512$ resolution. Lighting is from the
sampled HDRI. Each viewpoint generates two outputs: an RGB rendering and a 3D Box rendering. The latter features a directional 3D bounding box overlay, where the front, back, top, bottom, left, and right faces are color-coded in red, green, blue, yellow, pink, and cyan, respectively. Additionally, the 3D box rendering includes a depth-aligned planar floor rendered with a checkerboard pattern and depth-aware shading cues.

\begin{figure}[t]
    \centering
    \includegraphics[width=0.8\textwidth]{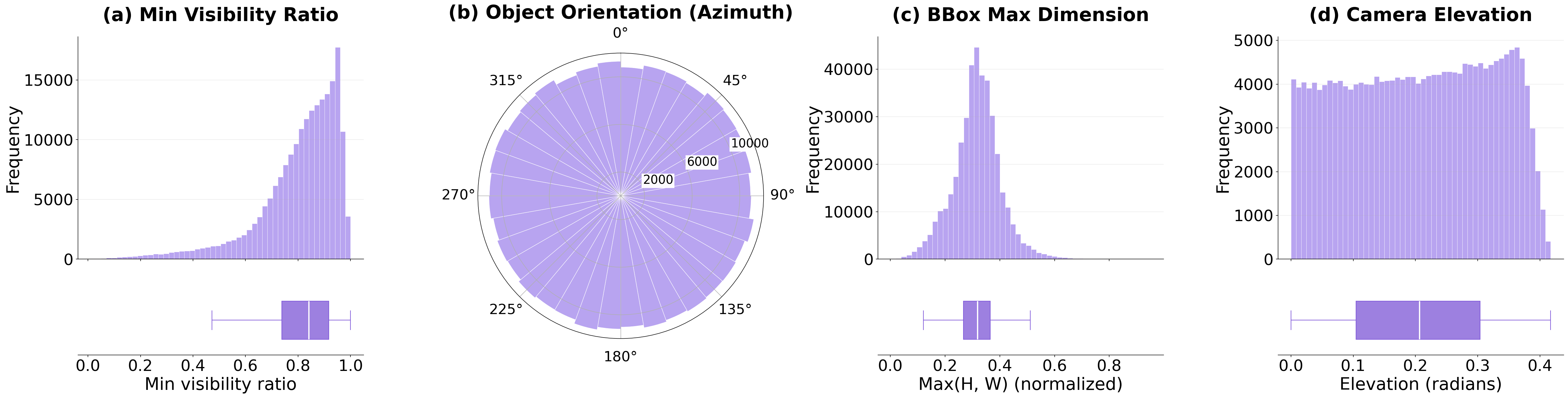}
    \caption{\textbf{Statistics of training dataset.}
    (a) The distribution of the minimum visibility ratio across objects within each scene is left-skewed and highly concentrated near unity (median=0.84). This profile indicates that while the majority of scenes exhibit only mild inter-object occlusion, a substantial long tail extending to 0.01 ensures the inclusion of severe occlusion cases.
    (b) Object orientations follow a near-uniform distribution over azimuth, avoiding any directional bias in the training data.
    (c) The maximum 2D bounding-box dimension (normalized by image side) is tightly concentrated around $0.32$, reflecting a deliberate scene composition that keeps multiple objects simultaneously visible at moderate scale rather than dominating the frame.
    (d) Camera elevation is broadly uniform from $0$ to roughly $0.4$ radians ($\approx 23^\circ$), this favors low-angle viewpoints over bird's-eye configurations, which in turn produce stronger inter-object occlusions and more visually challenging compositions.}
    \label{fig:dataset_stats}
\end{figure}

\begin{figure}[h]
    \centering
    \includegraphics[width=0.8\linewidth]{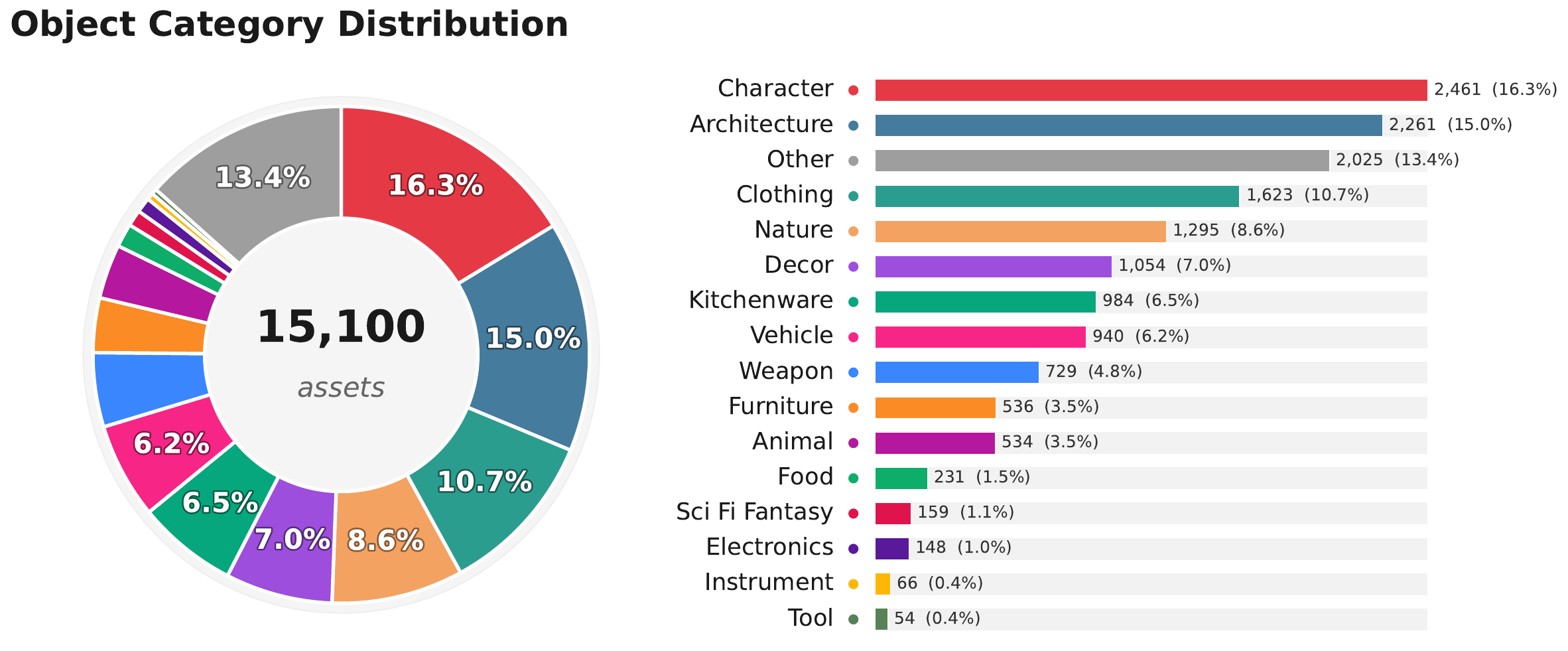}
    \caption{Object category distribution across the 10{,}143 downloaded assets, derived via multi-label keyword classification on the asset captions. \textit{Left:} donut chart visualizing the relative proportions of each category, with percentages shown for the dominant slices. \textit{Right:} ranked breakdown of all 16 categories with absolute counts and percentages. As assets may match multiple categories (e.g., a knight \textit{character} holding a \textit{weapon}), category counts sum to more than the total number of assets.}
    \label{fig:object_categories}
\end{figure}

\begin{figure}[H]
    \centering
    \includegraphics[width=0.8\linewidth]{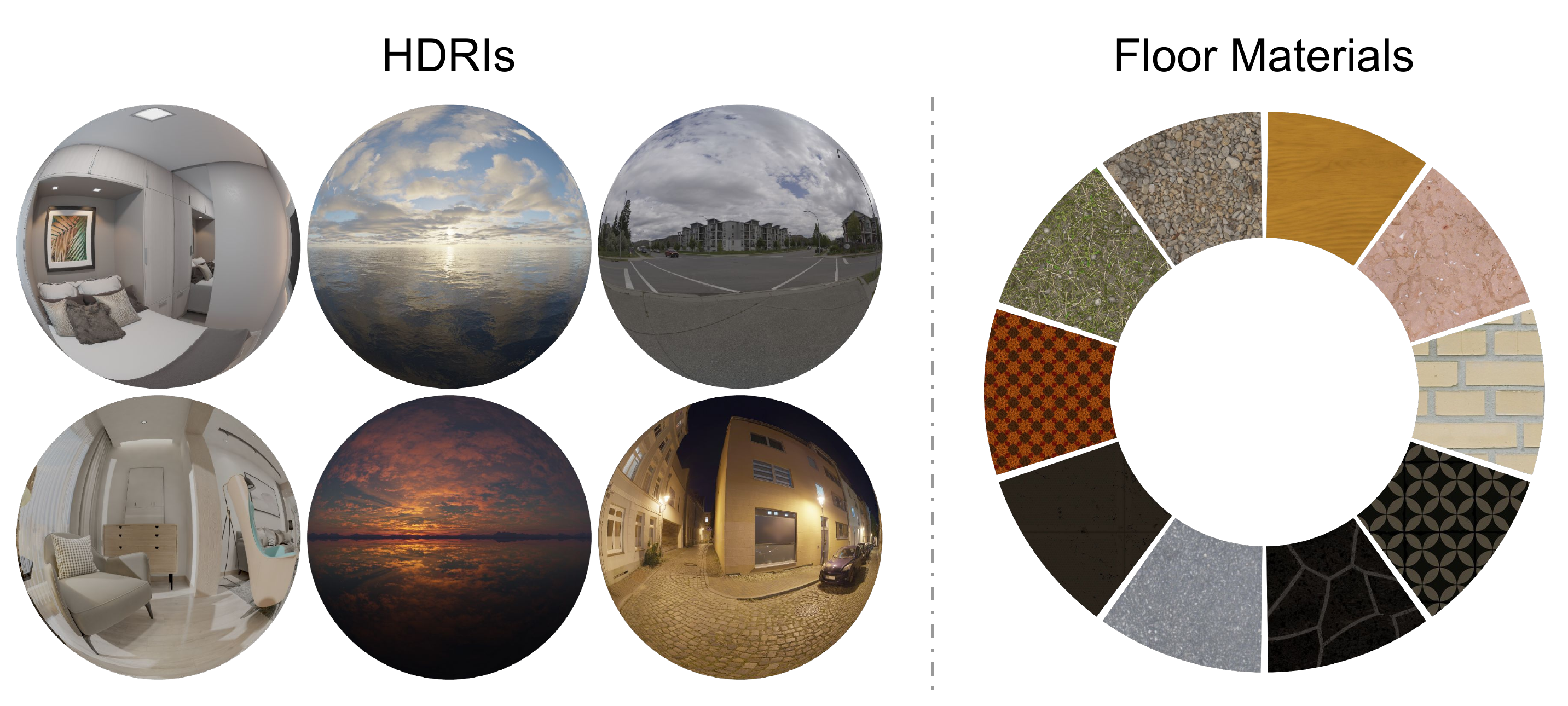}
    \caption{Sample illustrations from our dataset's environmental and material assets. \textit{Left:} Six representative HDRIs spanning indoor (top-left, bottom-left), natural outdoor (top-middle, bottom-middle), and urban outdoor (top-right, bottom-right) environments. \textit{Right:} A radial arrangement of 12 floor material samples drawn from our texture library, illustrating the diversity of surface categories (wood, stone, brick, tile, concrete, metal, etc.).}
    \label{fig:dataset_samples}
\end{figure}

\begin{figure}[H]
    \centering
    \includegraphics[width=\linewidth]{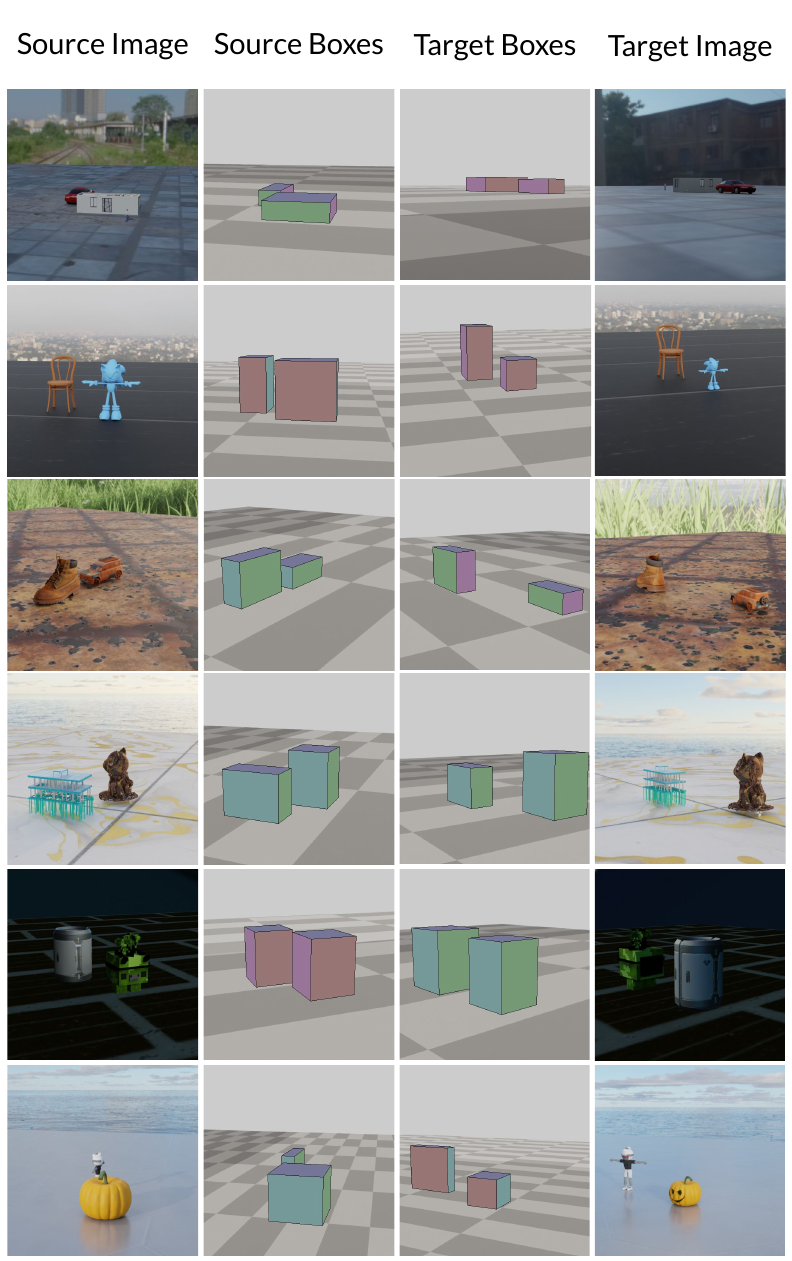}
    \vspace{-10pt}
    \caption{Various ObjaverseXL assets rendered in Blender, which are used in training are visualized in the above figure. The diversity of the lighting and environment condition are visualized.}
    \label{fig:supp_syndata_vis}
\end{figure}

Fig.~\ref{fig:dataset_stats} summarizes the distribution of our training data.  
Fig.~\ref{fig:dataset_samples} presents representative samples of the HDRI environments and floor materials used during rendering. 
The category distribution of the 10{,}143 objects sourced from the SketchFab subset of \textbf{ObjaverseXL}~\cite{objaverse-xl} is shown in Fig.~\ref{fig:object_categories}, and additional examples of our synthetic data rendered in Blender are provided in Fig.~\ref{fig:supp_syndata_vis}. Together, the wide variety of HDRIs, materials, and object categories yields a diverse training distribution that promotes generalization.

\section{Implementation details}
\label{sec:supp:impln}

We employ Flux-Kontext~\cite{flux-kontext} as our foundational image editing model.To integrate our specific conditioning, we train a LoRA adapter of rank 32, applied to the query, key, and value projection matrices across all MMDiT blocks. Throughout the editing process, we utilize an empty text prompt (""), ensuring the model relies exclusively on visual cues derived from our box conditioning.

The training process is executed in two distinct stages:

\begin{itemize}[leftmargin=*]
    \item \textbf{Stage 1:} The model is trained on a synthetic dataset of 100,000 scenes for 50,000 steps. This phase adapts the model’s internal representations to execute transformations dictated by the spatial relationship between source and target boxes.
     \item \textbf{Stage 2:} We fine-tune the model for 10,000 steps using a composite dataset of 20,000 scenes. This includes 10,000 samples from the Objectron~\cite{objectron} training set to improve performance on real-world imagery and 10,000 additional synthetic scenes. We retain synthetic data in this stage to ensure the model maintains robust object-level control, as the Objectron dataset focuses primarily on camera-motion-driven changes.
\end{itemize}

We utilize the Prodigy~\cite{prodigy} optimizer with safeguard warmup and bias correction enabled, maintaining a weight decay of 0.01. Training is conducted at a resolution of 512 × 512 with an effective batch size of 16. Our implementation is built on PyTorch~\cite{paszke2019pytorch} and the Hugging Face Diffusers~\cite{diffusers} library.

\newpage
\section{Ablation Visualizations}
\label{sec:supp:ablations}

\begin{figure}
    \centering
    \includegraphics[width=0.8\linewidth]{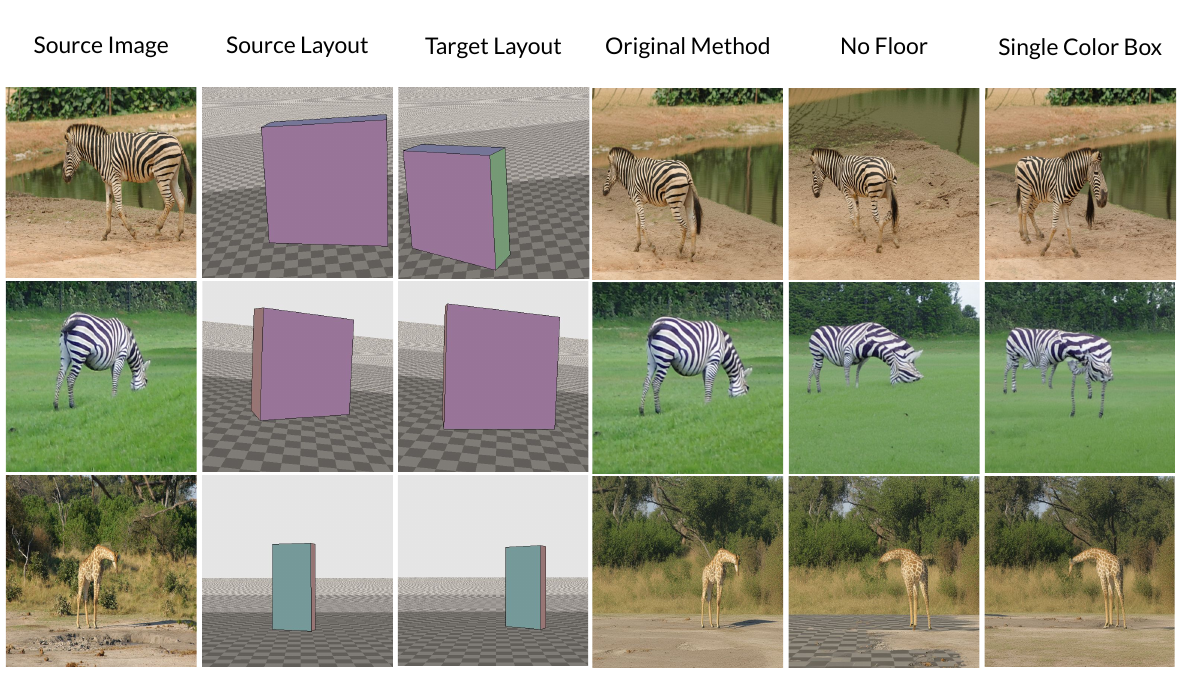}
\vspace{-5pt}
    \caption{\textbf{Ablation visualization.} 
Adding the checkered floor pattern for ground localization and orientation-specific face colors for the conditioning box yields generations that are substantially more faithful to the edit instructions. 
Without the floor (\textit{No Floor}), the model fails to preserve the texture and ground contact of the animals, as seen most clearly in the first row. 
With a uniformly colored box (\textit{Single Color Box}), the model loses the directional cue and frequently flips the animal's orientation to face the opposite direction. 
These results highlight the importance of encoding both ground reference and orientation information into the spatial conditioning signal.}
    \label{fig:suppl-quals-fig.3}
\end{figure}

\begin{figure}[H]
    \centering
    \includegraphics[width=0.8\linewidth]{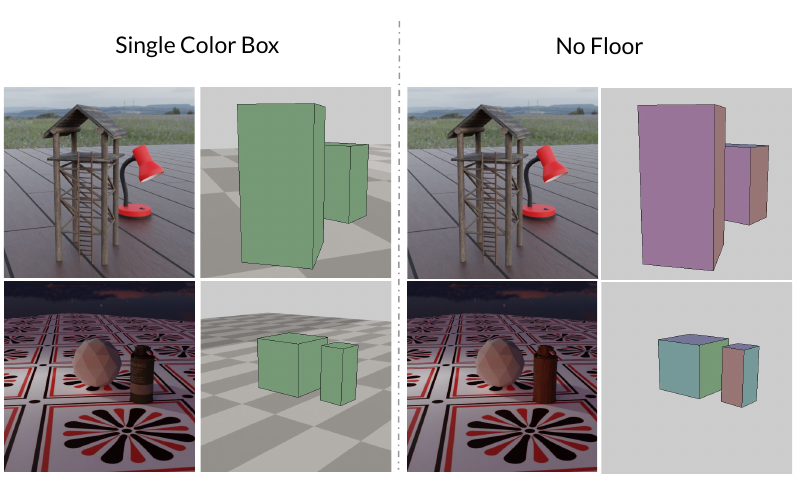}
\vspace{-5pt}
    \caption{\textbf{Ablation Dataset visualization.} 
Here we show a sample rendering of the assets with the same box colors and without floor.}
    \label{fig:suppl-ablation-samples}
\end{figure}

We provide qualitative examples in Fig.~\ref{fig:suppl-quals-fig.3} to complement the quantitative ablation results reported in the main paper. Each row corresponds to a different test sample, with columns showing the image generated with the original method, the \textit{No Floor} variant, and the \textit{Single Color Box} variant. Without the checkered floor, the model loses its global ground reference and produces outputs in which the animal's texture and contact with the ground are inconsistent with the target layout (most visible in the middle row). Without directional face colors, the model retains correct positioning but frequently flips the animal's facing direction, confirming that the directional coloring is the dominant cue for orientation. A sample of these box renderings are shown in Fig~\ref{fig:suppl-ablation-samples} Together, these examples visually reinforce the conclusion that the floor and directional box colors contribute complementary, non-redundant signals.

\section{User Study Setup}
\label{sec:supp:user-study}

\begin{figure}[H]
    \centering
    \includegraphics[width=0.8\linewidth]{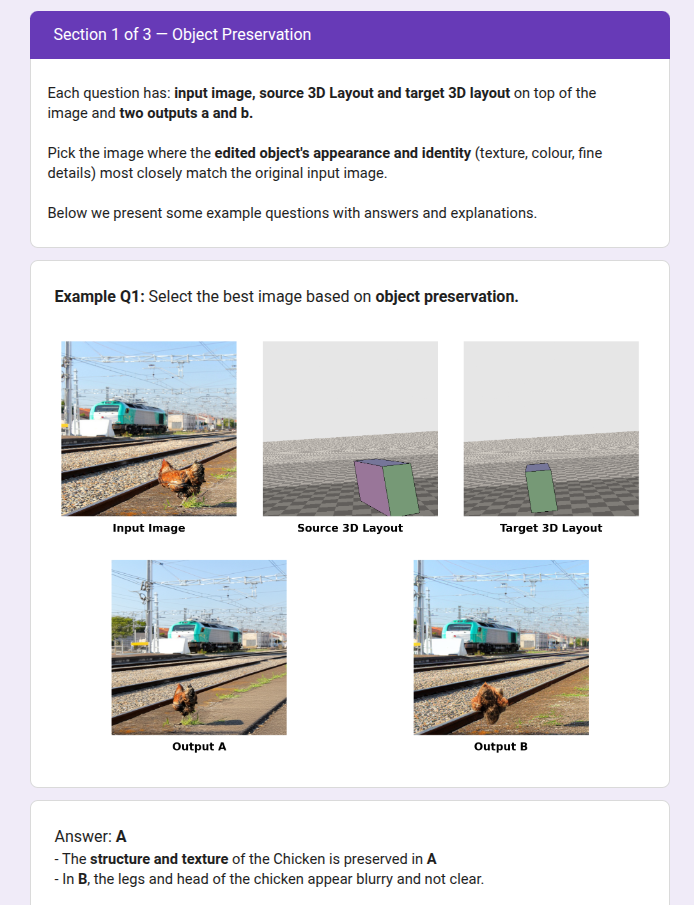}
    \caption{Participants are shown the input image, source and target 3D layouts, and two candidate edited outputs (A and B). They are asked to select the output that best preserves the edited object’s appearance and identity (e.g., foreground, background details) with respect to the original input image. Example questions with explanations are provided to familiarize participants with the evaluation protocol.}
    \label{fig:user_study_form}
\end{figure}

Fig.~\ref{fig:user_study_form} shows the interface used in our A/B user study. Each question presents the input image, the source and target 3D layouts, and two candidate edits, and participants select the preferred output along one of three criteria: \textit{object preservation} (texture, color, and fine details of the edited object), \textit{background preservation} (consistency of the surrounding scene), and \textit{layout following} (agreement with the target 3D layout). Example questions with reference answers are shown at the beginning of the study to familiarize participants with the evaluation protocol.

%%%%%%%%%%%%%%%%%%%%%%%%%%%%%%%%%%%%%%%%%%%%%%%%%%%%%%%%%%%%

\end{document}